\documentclass[]{gtech}
\usepackage{amssymb}
\usepackage{multirow}
\usepackage{bigdelim}
\usepackage{todonotes}
\usepackage{longtable}
\usepackage{tabularray}
\usepackage{wrapfig}
\usepackage[most]{tcolorbox} 
\usepackage{xcolor}
\usepackage{url}
\usepackage{colortbl}
\usepackage{amssymb}
\usepackage{pifont}
\usepackage{booktabs,multirow}
\usepackage{makecell}
\usepackage{tabulary}
\usepackage{fontawesome5}
\usepackage{bbding}
\usepackage{multicol}
\usepackage{amsmath}
\usepackage{subcaption}
\usepackage{graphicx}
\usepackage{amsfonts}
\usepackage{nicefrac}
\usepackage{microtype}

\usepackage{epigraph}
\newcommand{\se}{s_e}

\usepackage{algpseudocode}
\usepackage{amssymb}
\usepackage{mathtools}
\usepackage{amsthm}
\usepackage{algorithm}
\theoremstyle{plain}

\theoremstyle{definition}

\theoremstyle{remark}

\usepackage{bm}

\usepackage{tabularx}
\usepackage{enumitem}
\usepackage{threeparttable}

\usepackage{fancyhdr}
\usepackage{lastpage} 

\definecolor{mygray}{gray}{0.85}
\definecolor{LightCyan}{rgb}{0.88,1,1}
\setlist[itemize]{topsep=0pt,itemsep=0.2ex,parsep=0pt,partopsep=0pt,leftmargin=*,label=\textbullet}

\title{\textcolor[HTML]{2a7a1b}{Memory} Intelligence Agent}

\author{Jingyang Qiao\textsuperscript{$*$}, Weicheng Meng\textsuperscript{$*$}, Yu Cheng, Zhihang Lin, Zhizhong Zhang\textsuperscript{$\dagger$}, \\ Xin Tan, Jingyu Gong, Kun Shao, Yuan Xie\textsuperscript{$\ddagger$}}

\affiliation{East China Normal University, Shanghai Innovation Institute, Harbin Institute of Technology, Xiamen University, Shanghai Artificial Intelligence Laboratory, Independent Researcher}

\contribution{$*$ Equal Contribution $\dagger$ Corresponding Author $\ddagger$ Project Leader}

\abstract{\fontsize{11pt}{12pt} \textit{Deep research agents (DRAs) integrate LLM reasoning with external tools. Memory systems enable DRAs to leverage historical experiences, which are essential for efficient reasoning and autonomous evolution. Existing methods rely on retrieving similar trajectories from memory to aid reasoning, while suffering from key limitations of ineffective memory evolution and increasing storage and retrieval costs. To address these problems, we propose a novel \textbf{\textit{M}}emory \textbf{\textit{I}}ntelligence \textbf{\textit{A}}gent (MIA) framework, consisting of a Manager-Planner-Executor architecture. Memory Manager is a non-parametric memory system that can store compressed historical search trajectories. Planner is a parametric memory agent that can produce search plans for questions. Executor is another agent that can search and analyze information guided by the search plan. To build the MIA framework, we first adopt an alternating reinforcement learning paradigm to enhance cooperation between the Planner and the Executor. Furthermore, we enable the Planner to continuously evolve during test-time learning, with updates performed on-the-fly alongside inference without interrupting the reasoning process. Additionally, we establish a bidirectional conversion loop between parametric and non-parametric memories to achieve efficient memory evolution. Finally, we incorporate a reflection and an unsupervised judgment mechanisms to boost reasoning and self-evolution in the open world. Extensive experiments across eleven benchmarks demonstrate the superiority of MIA. First, MIA significantly enhances the current SOTA LLMs' performance in deep research tasks. For instance, MIA further boosts GPT-5.4 performance by up to 9\% and 6\% on LiveVQA and HotpotQA, respectively. Furthermore, with the lightweight Executor, like Qwen2.5-VL-7B, MIA can also achieve an average improvement of 31\% across evaluated datasets, outperforming the much larger Qwen2.5-VL-32B by a margin of 18\%, highlighting its remarkable performance. Additionally, training analysis reveals that reinforcement learning enables the Planner and Executor to synergistically optimize their strategies, effectively capturing dataset-specific characteristics and enhancing cross-domain reasoning and memory capabilities. Tool analysis reveals that long-context memory methods struggle with multi-turn tool interaction, while our proposed MIA significantly outperforms previous methods. Under unsupervised settings, MIA achieves performance comparable to its supervised counterpart, meanwhile exhibiting the progressive self-evolution performance across multiple training iterations.}}

\gtechdata[Code]{\url{https://github.com/ECNU-SII/MIA}}
\gtechdata[Model]{\url{https://huggingface.co/LightningCreeper/MIA}}
\gtechdata[Dataset]{\url{https://huggingface.co/datasets/LightningCreeper/MIA}}

\begin{document}
\maketitle

\begin{figure*}[h]
    \centering
    \includegraphics[width=0.93\textwidth]{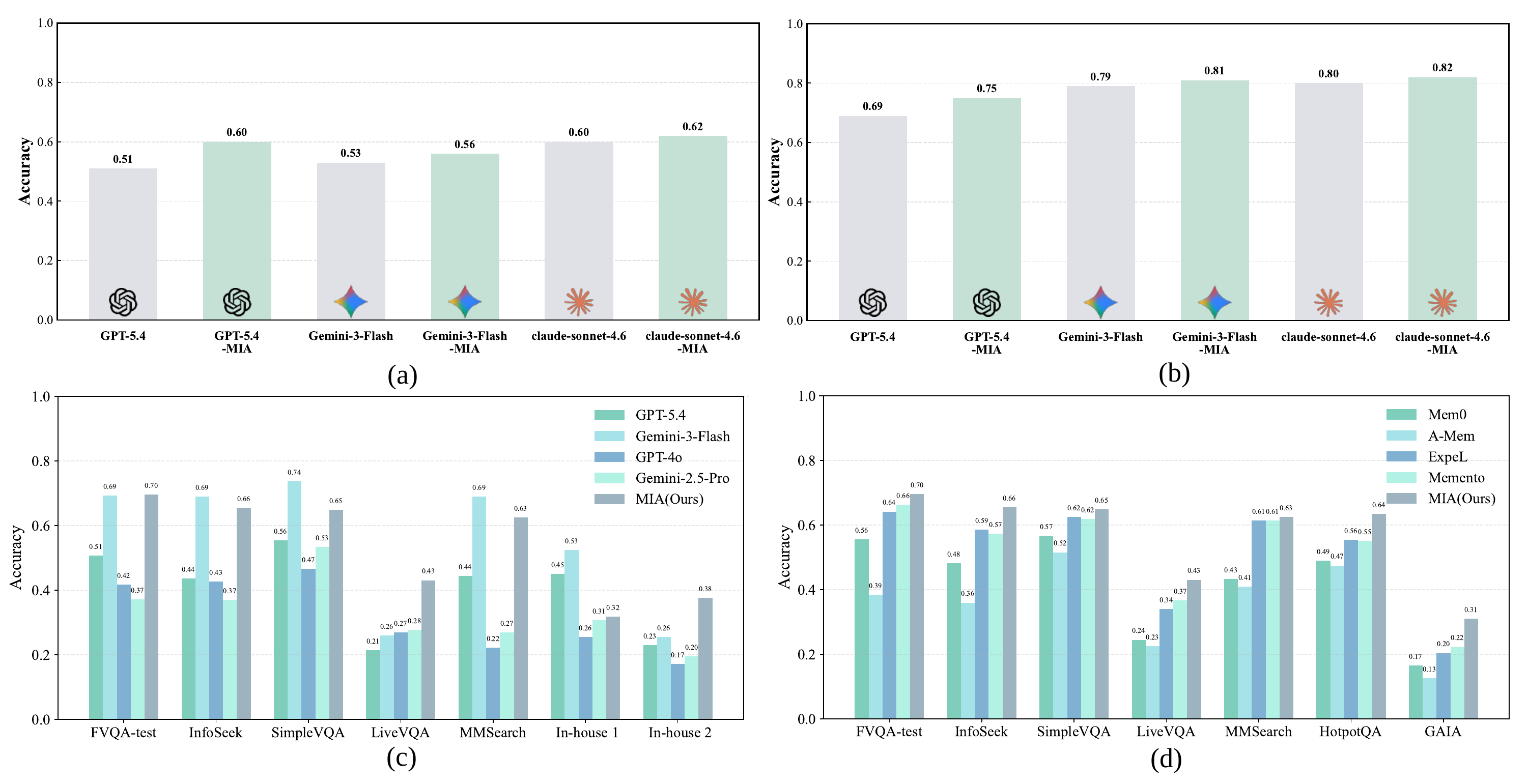}
    \caption{(a). Comparisons between frontier LLMs and their MIA-enhanced counterparts on the LiveVQA (multimodal) dataset. (b). Comparisons between frontier LLMs and their MIA-enhanced counterparts on the HotpotQA (text-only) dataset. (c). Comparisons between MIA based on Qwen2.5-VL-7B Executor with larger LLMs (in non-tool-calling settings) across seven diverse datasets. (d). Comparisons between MIA and SOTA memory frameworks based on Qwen-2.5-VL-7B Executor across seven diverse datasets. }
    \vspace{-3mm}
    \label{fig}
\end{figure*}

\vspace{3mm}
\epigraph{"\textit{Never memorize something that you can look up}."}{---Albert Einstein} 

\section{Introduction}
\label{sec:intro}
Deep Research Agents (DRAs)~\citep{survey1,survey2,survey3} can combine the reasoning capabilities of LLMs with external tools, such as search engines, thereby empowering LLMs to complete complex, open-ended tasks. Based on tool-augmented LLMs~\citep{agentlab,talm,api,sciagent}, DRAs follow a multi-round paradigm with repeatedly interleaved reasoning and external searching~\citep{webthinker,deepresearchbench,benchmarking}. As agents evolve toward long-horizon, multi-turn interactions, memory systems become a critical component~\citep{agentmem,multiagentmem,mirix}. They determine whether the agent can accumulate experience, refine search strategies, and improve during each research process rather than repeatedly solving each task from scratch~\citep{memsurvey1,multiagent}. Existing research on agent memory has mainly focused on long context scenarios~\citep{gmem,zep}, where information is stored based on the traces of search experience. Although such approaches have shown promising performance in many agentic applications~\citep{infllm}, long-context memory exhibits fundamental limitations when applied to deep research agents~\citep{long,look,longmemeval}. First, long contexts may dilute attention, hindering the model’s understanding of the current problem. Second, irrelevant or weakly related content in memory introduces noise, leading to degraded reasoning ability. Third, maintaining ever-growing context histories poses substantial storage challenges, particularly for agents operating continuously over extended periods. Finally, retrieval over massive memory incurs increasing computational costs, resulting in time inefficiency.

Furthermore, long-context memory primarily captures knowledge-oriented or factual-oriented memory describing what the result is (\textit{e.g.}, user attributes, historical facts, and retrieved documents)~\citep{augmenting,memagent,lm2}. In contrast, deep research relies heavily on process-oriented memory~\citep{memp} and conceptual knowledge describing how a result is obtained (\textit{e.g.}, search trajectories, failed attempts, and successful reasoning strategies). The objective of adopting memory is not merely to store retrieved knowledge, but to leverage historical experiences to guide future planning and exploration~\citep{hiagent,remember}. Therefore, deep research agents require memory mechanisms that assist in search path planning and strategy reuse, rather than simply expanding the amount of stored textual context.

To address the limitations of long-context memory applied in deep research agents, existing memory systems typically utilize pre-trained models as planners to generate chain-of-thought (CoT) prompting for search path planning with few-shot cases~\citep{memento}. While such methods have improved the deep research performance, they still suffer from several key challenges: (1) The Planner operates without task-specific training, resulting in suboptimal planning. (2) Previous CoT-based prompting methods select few-shot examples only based on relevance, while neglecting quality, frequency, and other significant dimensions. (3) The Executor fails to adequately interpret and follow planning instructions without task-specific training. In summary, the essence of prior works can be characterized as an incompetent Planner retrieving memories from bloated memory and using non-comprehensive in-context prompts to guide an unprepared Executor in conducting deep research. Consequently, introducing memory systems yields limited performance improvements.

To address these challenges, we propose the Memory Intelligence Agent (MIA), a novel framework that integrates brain-inspired memory mechanisms into a Manager-Planner-Executor architecture. Specifically, MIA employs a hippocampus-like episodic memory to extract insights from historical trajectories. Meanwhile, it consolidates historical trajectories into parametric memory via Planner training, reducing storage overhead. Then, it trains the Executor to follow and execute the generated plan, enabling synergistic co-evolution between the two agents. Finally, it introduces a reflection mechanism to develop the autonomous re-planning ability, paving the way for self-evolution under sparse annotations or unsupervised conditions. Extensive experiments demonstrate that (1) MIA significantly elevates the performance of state-of-the-art (SOTA) Executors. Specifically, it yields a 9\% improvement on the LiveVQA benchmark and a 6\% gain on HotpotQA when integrated with GPT-5.4, showcasing its ability to further enhance even the most powerful models. (2) MIA exhibits remarkable improvements for smaller Executors. Using Qwen2.5-VL-7B as the Executor, our framework achieves an average gain of 31\% across seven diverse datasets, notably outperforming its much larger counterpart, Qwen2.5-VL-32B, by 18\%. (3) Under unsupervised settings, MIA empowers the trained Executor to achieve a 7\% performance boost. Furthermore, we observe consistent performance growth over multiple training iterations, validating the effectiveness of our autonomous evolution mechanism. (4) MIA sets a new state-of-the-art. Building on the Qwen2.5-VL-7B Executor, our approach consistently outperforms previous SOTA memory baselines by an average margin of 5\% across all seven evaluated benchmarks. 

Our contributions are as follows:

\textbullet{} We introduce a Manager-Planner-Executor architecture that addresses the storage bottlenecks and reasoning inefficiencies of conventional deep research agents by decoupling of historic memory, parametric planning and dynamic execution.

\textbullet{} We propose an alternating RL paradigm to optimize the interplay between the Planner and Executor. This ensures that high-level planning and low-level retrieval are mutually aligned.

\textbullet{} We develop a continual test-time learning mechanism, allowing the Planner to update its parametric knowledge during inference. This enables the agent to adapt to new information without interrupting the reasoning workflow.

\textbullet{} We integrate reflection and unsupervised judgment mechanisms, endowing the agent with self-assessment and correction capabilities in open-ended tasks. This not only enhances reasoning robustness but also ensures continual evolution when facing unknown tasks.

\textbullet{} MIA surpasses existing memory baselines and exhibits strong scalability, significantly enhancing the performance of both frontier and small-scale LLMs in deep research tasks.

\section{Related Work}

\begin{figure*}[t]
    \vspace{-3mm}
    \centering
    \includegraphics[width=0.89\textwidth]{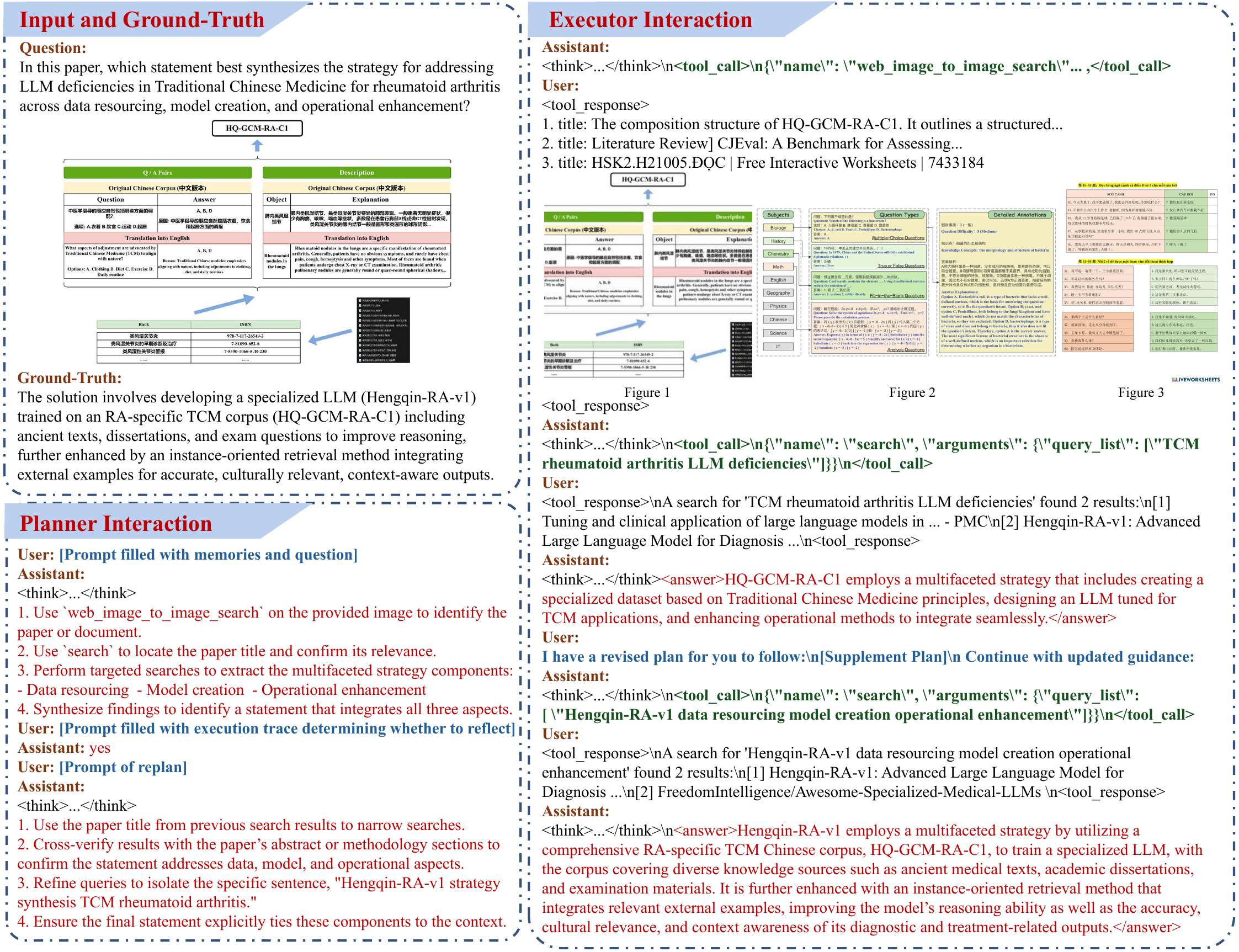}
    \caption{A deep research process of MIA to tackle a complex and multi-hop question.}
    \vspace{-5mm}
    \label{fig_case}
\end{figure*}

\label{sec:rel_work}
\textbf{Deep Research Agents:} handle complex and dynamic tasks by iteratively prompting LLMs to perform external search and reason over retrieved results. However, without parameter optimization through the training process, LLMs fail to grasp effective tool calling and adapt to real-world environments. Recent methods, such as DeepResearcher~\citep{deepresearcher} and Search-R1~\citep{searchr1} have used reinforcement learning (RL) to enhance multi-turn search and retrieval-augmented generation, while they are confined to text-only tasks. Based on these foundations, MMSearch-R1~\citep{mmsearchr1}, and DeepMMSearch-R1~\citep{deepmmsearchr1} further integrate multimodal search tools and significantly improves reasoning in multimodal search. Despite progress having been made in the integration of external retrieval and internal reasoning, challenges such as low efficiency in utilizing historical search information and past experiences still exist.

\textbf{Agent Memory Systems:} ReasoningBank~\citep{reasoningbank} and MemoryBank~\citep{memorybank} enhance the agent’s reasoning ability through the scalable expansion of memory. The ExpeL~\citep{expel} framework optimizes decision-making by learning from successful and failed experiences. Mem-$\alpha$~\citep{memalpha} and Memory-r1~\citep{memr1} incorporate RL to model memory as a Markov decision process and guide agents to learn optimal memory storage strategies through reward signals. In memory management, Agentic Memory~\citep{agenticmem} and A-Mem~\citep{amem} propose long-term and short-term memory and graph-based memory management paradigms, respectively, significantly improving the agent's contextual scheduling capabilities in complex tasks. Memento~\citep{memento} proposes a different approach by exploring performance gains through memory fine-tuning while freezing the LLMs. In memory evolution, MemEvolve~\citep{memevolve} drives dynamic adjustments of memory systems through higher-order meta-feedback, while Evo-Memory~\citep{evomem} constructs benchmarks to evaluate the autonomous evolution capability of memory systems during inference. Despite progress having been made in the construction and management of memory systems, challenges such as low memory efficiency, instability of results, and poor interpretability still exist. Additionally, autonomous evolution for memory systems in unsupervised environments is unexplored.

\begin{figure*}[t]
    \vspace{-3mm}
    \centering
    \includegraphics[width=0.85\textwidth]{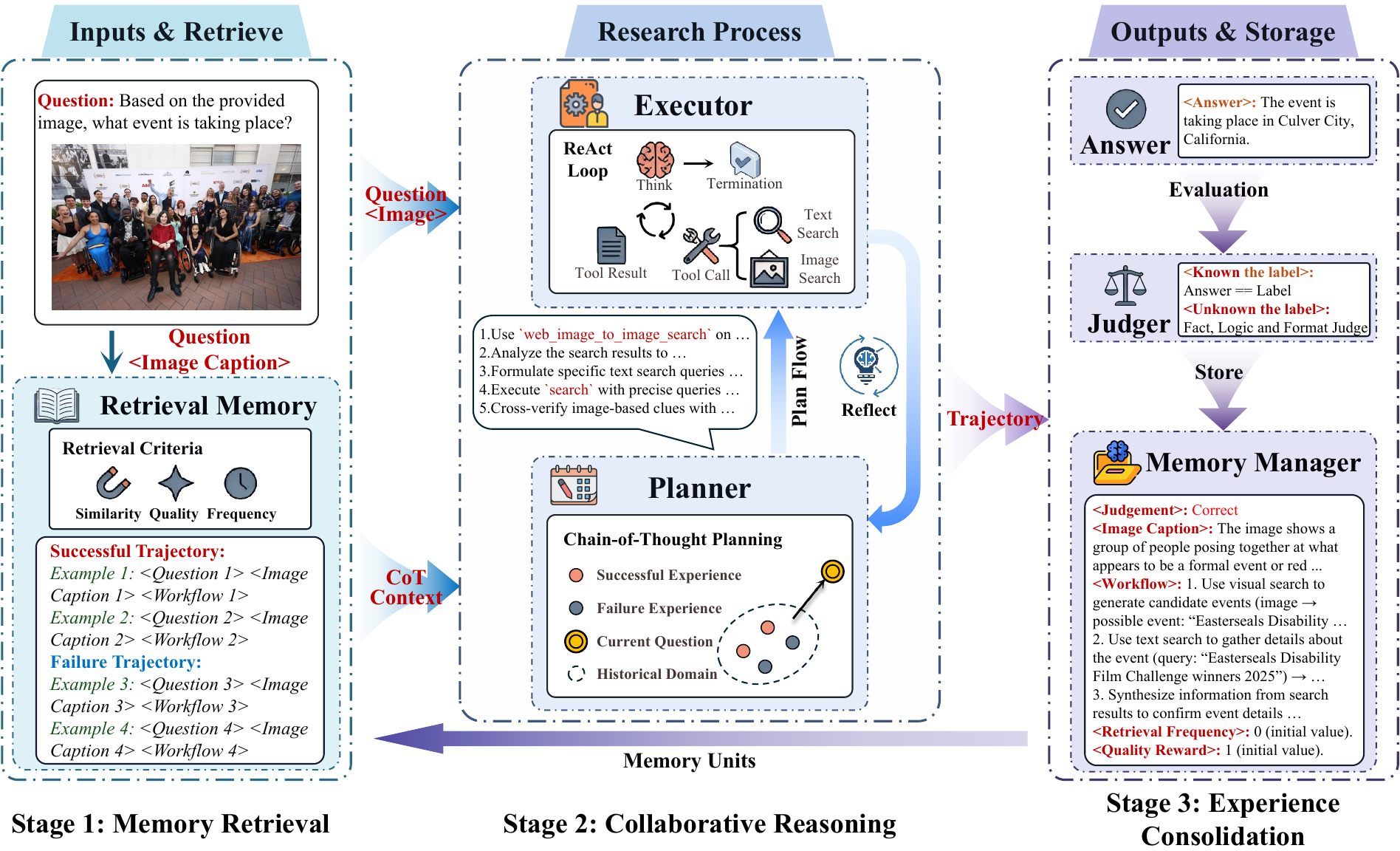}
    \caption{Reasoning process of MIA consists of three parts: Inputs \& Retrieval is for retrieving memory context similar to the inputs; Research Process is for driving Planner-Executor collaboration via a planning-execution-reflection loop; Outputs \& Storage is for compressing search trajectories into structured memory.}
    \vspace{-5mm}
    \label{fig_1}
\end{figure*}

\section{Method}
\label{sec:method}
\subsection{Overview} 
Figure~\ref{fig_1} illustrates the architecture and reasoning process of the proposed MIA. In the architecture, the Planner is an agent for generating a search plan for questions. The Executor is responsible for implementing the plan step by step until obtaining the final result. These two agents are initialized from pre-trained LLM (\textit{e.g.}, Qwen3-8B~\citep{qwen3}) and LMM (\textit{e.g.}, Qwen2.5-VL-7B~\citep{qwen25vl}). The Memory Manager is a system composed of a memory buffer and a pre-trained LLM (\textit{e.g.}, Qwen3-32B~\citep{qwen3}). The memory buffer is responsible for saving high-value historical trajectories which serve as CoT cases. The pre-trained LLM is frozen and served to manage the buffer with context prompts. In the reasoning process, the Planner first analyzes historical cases to formulate a search plan for the current question. After that, the Executor interprets this plan, performing task reasoning in conjunction with tool usage, and provides feedback to the Planner to trigger reflection and determine the necessity of re-planning. Finally, the reasoning results and execution processes are submitted to the Memory Manager for further memory compression and structured organization. 

Figure~\ref{fig_2} depicts the memory framework of MIA, which enables the agent's lifelong learning through two complementary mechanisms: non-parametric memory for contrastive experience and parametric memory for long-term self-consolidation. Specifically, the non-parametric memory retrieves similar and high-value historical records to provide an explicit reference for the Planner. Meanwhile, the parametric memory distills latent knowledge representations from trajectories, internalizing them into the agent's intrinsic capabilities to achieve evolution.

\subsection{MIA Agent Loop}
MIA is equipped with a planning-execution-memory loop for the agent's lifelong learning. As shown in Figure~\ref{fig_1}, the agent loop comprises three main stages: memory retrieval, collaborative reasoning, and experience consolidation. In summary, our objective is to enable the frozen agent to achieve lifelong evolution in a continuous data stream by leveraging MIA.

\subsubsection{Memory Retrieval}
The deep research process initiates with the parsing of a multimodal query. At the beginning, the Memory Manager is empty, and memory retrieval is not employed. Once a sufficient amount of historical experiences (trajectories) have been accumulated, we retrieve similar past experience based on the current input. Specifically, the Memory Manager converts visual inputs into textual captions, aligning them with the storage format of the units in the Memory Manager. A hybrid retrieval strategy then scores each memory across the following three dimensions:
\begin{itemize}
    \item \textbf{Semantic Similarity:} Measures the semantic distance between the current input and the historical questions and image captions stored in the memory units to ensure contextual relevance.
    \item \textbf{Value Reward:} Prioritizes memory units with high success rates to provide high-quality context.
    \item \textbf{Frequency Reward:} Rewards low-frequency memory units to encourage the exploration of long-tail, potentially relevant knowledge, thereby maintaining a balance in memory utilization.
\end{itemize}
Based on these scores, MIA overcomes the limitations of conventional single-dimensional memory retrieval. The framework retrieves both successful trajectories (\textit{positive paradigms}) and failed trajectories (\textit{negative constraints}) to construct a rich contextual prior for planning. More details are provided in Appendix \ref{retrieval}.

\subsubsection{Collaborative Reasoning}
The core reasoning process is driven by the dynamic synergy between two trainable LLMs:
\begin{itemize}
    \item \textbf{Planner (Cognitive Hub):} With the retrieved trajectories, it adopts a few-shot CoT strategy to decompose complex tasks into executable step-by-step sub-goals and generates a plan.
    \item \textbf{Executor (Operational Terminal):} After planning, it interacts with the environment via a ReAct loop according to the plan. By using external search tools and reasoning on the search results, it gathers and analyzes information to derive the final answer.
\end{itemize}
Additionally, we design a dynamic feedback loop to connect the two agents: 1) The Executor reports execution status to the Planner after obtaining the final answer. 2) The Planner triggers \textbf{Reflect-Replan} mechanism to dynamically adjust the search plan conditioned on the execution feedback. For example, if the Executor encounters an impasse or unexpected environmental feedback, the Planner will generate a new plan, and the Executor will then proceed with the revised plan. To reduce reasoning time, the Reflect-Replan mechanism is triggered only once.

\begin{figure*}[t]
    \vspace{-3mm}
    \centering
    \includegraphics[width=1.0\textwidth]{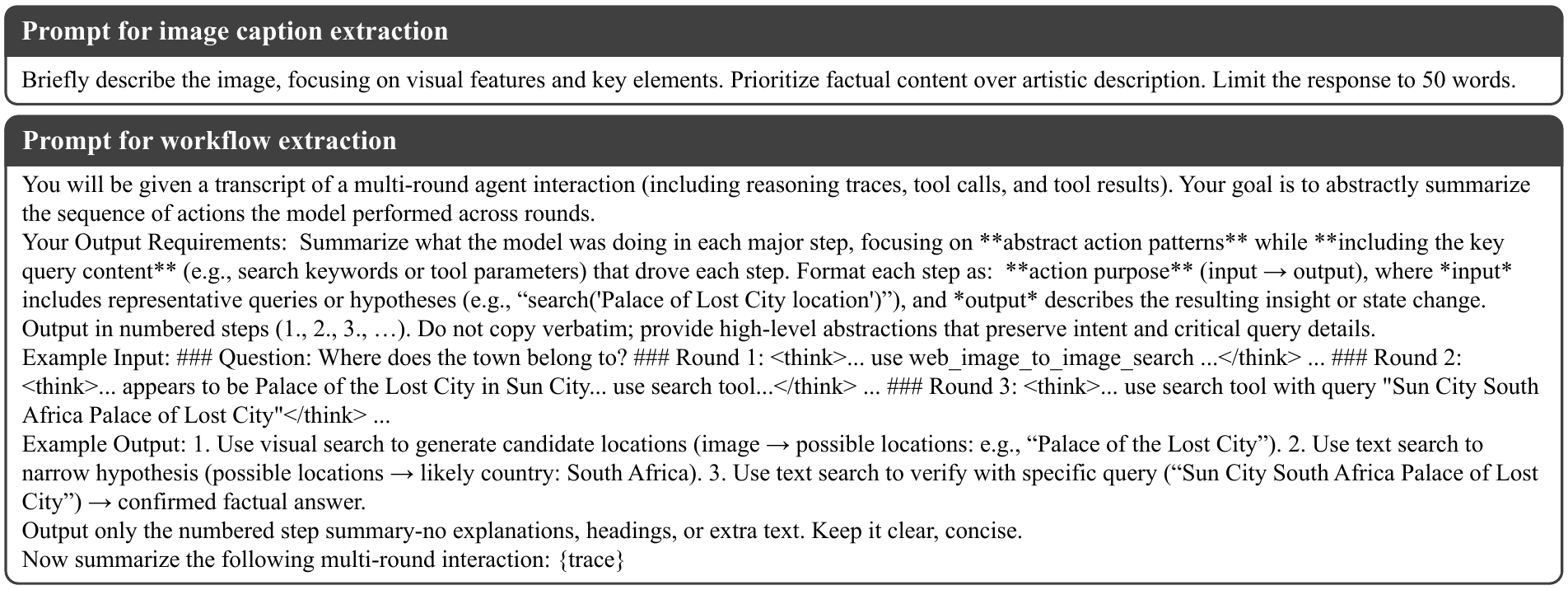}
    \vspace{-7mm}
    \caption{Prompt for Memory Manager to extract image caption and workflow.}
    \vspace{-3mm}
    \label{fig_prompt1}
\end{figure*}

\subsubsection{Experience Consolidation}
An LLM Judger is utilized to evaluate the final result. The Manager then compresses the current trajectories and updates the memory units. First, to reduce storage and retrieval burdens, images are compressed into captions, and verbose trajectories are abstracted into structured workflow summaries to form the new memory, as shown in Figure~\ref{fig_prompt1}. A Qwen3-32B LLM is utilized to complete this compression task. After that, we calculate the semantic similarity between the new memory and existing memory units to conduct a high semantic similarity knowledge replacement. If no similar memory unit is found, the new memory is stored as a new memory unit. Concurrently, the value rewards and frequency counts of relevant units are also updated. Finally, we propose a real-time exploration and update strategy to retrain the Planner with the current batch of questions, trajectories, and results, thereby internalizing episodic memory into parametric memory. This strategy enables further memory compression and efficient memory extraction. After training, the memory units will be selectively cleared. This prevents memory explosion while retaining critical information.

\subsection{Two-Stage Alternating RL Training}
Aiming to enhance the collaborative reasoning and promote the connection between the Planner and the Executor, we propose a two-stage alternating RL training strategy based on Group Relative Policy Optimization (GRPO)~\citep{grpo} in the training process. In stage 1, we focus on training the Executor to understand and follow the plan generated by the Planner, further improving the reasoning capability of the Executor. The Planner is frozen as a server to provide plans. After training in stage 1, we utilize the Planner and the trained Executor to collect training data containing memory contexts. In stage 2, we freeze the Executor and deploy it as a server. Based on the collected contexts, we conduct RL on the Planner to enhance its plan generation and reflection capabilities.

\subsubsection{Executor RL Training}
To endow the Executor with the capabilities of deep reasoning, tool calling under the ReAct~\citep{react} paradigm, as well as to enable it to accurately parse the Plan and Replan instructions generated by the Planner, we extend the GRPO objective function based on the interaction between tools and the Planner:
\begin{align}\label{eq:grpo_executor}
\mathcal{J}_{GRPO}^{M_E}(\theta) = \, & 
\mathbb{E}_{x \sim \mathcal{D}, \{ y_i \}_{i=1}^{G} \sim \pi_{\text{old}}( \cdot| x; \se, M_P)}\Bigg[\frac{1}{G} \sum_{i=1}^{G} \frac{1}{\sum_{t=1}^{|y_i|}  I(y_{i,t})} \sum_{t=1}^{|y_i|} \min \Bigg(\frac{\pi_{\theta}(y_{i,t} | x, y_{i,<t}; \se, M_P)}{\pi_{\text{old}}(y_{i,t} | x, y_{i,<t}; \se, M_P)} \nonumber \\ &\hat{A}_{i,t}, \text{clip} \Bigg( \frac{\pi_{\theta}(y_{i,t} | x, y_{i,<t}; \se, M_P)}{\pi_{\text{old}}(y_{i,t} | x, y_{i,<t}; \se, M_P)}, 1 - \epsilon, 1 + \epsilon \Bigg) \hat{A}_{i,t} \Bigg) -\beta \mathbb{D}_{KL} \left[ \pi_{\theta} || \pi_{\text{ref}} \right]\Bigg],
\end{align}
where $\pi_\theta$ and $\pi_{\text{old}}$ represent the current and previous policy models, $\pi_{\text{ref}}$ is the reference model. $\mathbb{D}_{KL}$ refers to the KL-divergence. $x$ denotes input samples drawn from the dataset $D$, including $\text{<question>}$ and $\text{<image>}$. $\epsilon$ and $\beta$ are hyperparameters. $\hat{A}_{i,t}$ indicates the advantage, computed based on the relative rewards of outputs within each group. $S_e$ refers to historical trajectories. $M_P$ is the Planner. $y$ expresses the interleaved trajectory generated by the interaction among the policy model, toolset, and Planner. $I(y_{i,t})$ is a token loss masking operation: $I(y_{i,t})=1$ if $y_{i,t}$ is generated by $\pi_\theta$, and $0$ if it is generated by tools or the Planner. The rollout process is shown in Table~\ref{tab:executor_training}.
\begin{table}[ht]
\centering
\small
\vspace{-1mm}
\caption{Executor Training Rollout Process.}
\label{tab:executor_training}
\begin{tabularx}{\linewidth}{l X}
\toprule
\textbf{Step} & \textbf{Action / Description} \\
\midrule
\textbf{Step 1} & Provide <question> to Planner to obtain the initial plan. $\triangleright$ \textbf{Step 2}. \\
\textbf{Step 2} & Input <question>, <image>, and initial plan to the policy model. $\triangleright$ \textbf{Step 3}. \\
\textbf{Step 3} & Generate thought(<think>) and action(<tool\_call> or <answer>) based on current state. If a tool call is needed $\triangleright$ \textbf{Step 4}; otherwise, generate candidate answer $\triangleright$ \textbf{Step 5}. \\
\textbf{Step 4} & Execute the corresponding tool and append the returned observation. $\triangleright$ \textbf{Step 3}. \\
\textbf{Step 5} & LLM Judger evaluates the candidate answer. If correct $\triangleright$ \textbf{Step 7}; otherwise $\triangleright$ \textbf{Step 6}. \\
\textbf{Step 6} & Provide interaction history to Planner to get a revised plan. $\triangleright$ \textbf{Step 3}. \textit{(Note: This action triggers at most once; if already triggered} $\triangleright$  \textbf{Step 7}). \\
\textbf{Step 7} & Output the final response. \\
\bottomrule
\vspace{-7mm}
\end{tabularx}
\end{table}

The reward function serves as the primary signal guiding the RL optimization. To train the Executor, we design a rule-based composite reward function:
\begin{gather}
\label{eq:reward_executor}
    r_{M_E}(x, y) = 0.7 * r_{1}(a_\text{pred}, a_\text{gold}) + 0.2 * r_{2}(y) + 0.1 * r_{3}(y),
\end{gather}
where $a_\text{pred}$ is the final answer extracted from the response $y$, and $a_\text{gold}$ is the ground truth. $r_{3}$ represents the format reward, which is $1$ if the output format complies with the specifications, and $0$ otherwise. $r_{2}$ denotes the tool reward, which is $1$ if there is a successful and standardized tool call, and $0$ otherwise. $r_{1}$ indicates the correctness reward evaluated by the LLM Judger, which is $1$ if the predicted answer $a_{\text{pred}}$ matches the ground truth $a_{\text{gold}}$, and $0$ otherwise.

\subsubsection{Planner RL Training}
Based on the trained Executor, which possesses instruction parsing and reasoning capabilities, stage 2 aims to optimize the Planner by incorporating memory contexts. The objective is to enhance the Planner's ability to absorb memory, plan for complex questions, and reflect on the feedback from the Executor. We extend the GRPO objective function based on the memory context and the Executor:
\begin{align}\label{eq:grpo_planner}
\mathcal{J}_{GRPO}^{M_P}(\theta) = \, & 
\mathbb{E}_{x \sim \mathcal{D}, \{ y_i \}_{i=1}^{G} \sim \pi_{\text{old}}( \cdot| m, x; M_E)}\Bigg[\frac{1}{G} \sum_{i=1}^{G} \frac{1}{\sum_{t=1}^{|y_i|}  I(y_{i,t})} \sum_{t=1}^{|y_i|} \min \Bigg(\frac{\pi_{\theta}(y_{i,t} | m, x, y_{i,<t}; M_E)}{\pi_{\text{old}}(y_{i,t} | m, x, y_{i,<t}; M_E)} \nonumber \\ & \hat{A}_{i,t}, \text{clip} \Bigg( \frac{\pi_{\theta}(y_{i,t} | m, x, y_{i,<t}; M_E)}{\pi_{\text{old}}(y_{i,t} | m, x, y_{i,<t}; M_E)}, 1 - \epsilon, 1 + \epsilon \Bigg) \hat{A}_{i,t} \Bigg) - \beta \mathbb{D}_{KL} \left[ \pi_{\theta} || \pi_{\text{ref}} \right]\Bigg],
\end{align}
where $m$ denotes the retrieved memory context. $M_E$ represents the Executor trained in stage 1. $y$ denotes the interleaved response trajectory generated by the interaction between the Planner and the Executor. The rollout process during the Planner training is shown in Table~\ref{tab:planner_training}.
\begin{table}[ht]
\centering
\small
\vspace{-1mm}
\caption{Planner Training Rollout Process.}
\label{tab:planner_training}
\begin{tabularx}{\linewidth}{l X}
\toprule
\textbf{Step} & \textbf{Action / Description} \\
\midrule
\textbf{Step 1} & Input the memory context, <question>, and prompt template to the policy model. $\triangleright$ \textbf{Step 2}. \\
\textbf{Step 2} & Perform rollout to generate initial plan through CoT. $\triangleright$ \textbf{Step 3}. \\
\textbf{Step 3} & Executor interacts with the environment using <question>, <image>, tools, and initial plan, yielding the candidate trajectory and result. $\triangleright$ \textbf{Step 4}. \\
\textbf{Step 4} & Analyze candidate trajectory. If interaction should terminate $\triangleright$ \textbf{Step 7}; otherwise $\triangleright$ \textbf{Step 5}. \\
\textbf{Step 5} & Perform rollout to generate CoT reflection and revised plan. $\triangleright$ \textbf{Step 6}. \\
\textbf{Step 6} & Executor continues interaction based on candidate trajectory and revised plan, yielding final trajectory and result. $\triangleright$ \textbf{Step 7}. \\
\textbf{Step 7} & Output the final response. \\
\bottomrule
\vspace{-7mm}
\end{tabularx}
\end{table}

The reward function for the Planner training comprises the following four dimensions:
\begin{gather}
\label{eq:reward_planner}
    r_{M_P}(x, y) = 0.7 * r_{1}(a_\text{pred}^{2}, a_\text{gold}) + 0.2 * r_{1}(a_\text{pred}^{1}, a_\text{gold}) + 0.05 * r_{2}(y, a_\text{gold}) + 0.05 * r_{3}(y),
\end{gather} where $a_\text{pred}^{1}$ is the intermediate answer, and $a_\text{pred}^{2}$ is the final answer. The two answers are equivalent if no reflection occurs. $r_{3}$ represents the format reward. $r_{2}$ indicates the reflection reward, which is designed to encourage an effective reflection mechanism (which is 1 if the first interaction is correct and reflection is uninitiated, or if the interaction is incorrect and reflection is initiated; otherwise, it is 0). $r_{1}$ refers to the correctness reward evaluated by the LLM Judger. 

\begin{figure*}[t]
    \centering
    \includegraphics[width=0.95\textwidth]{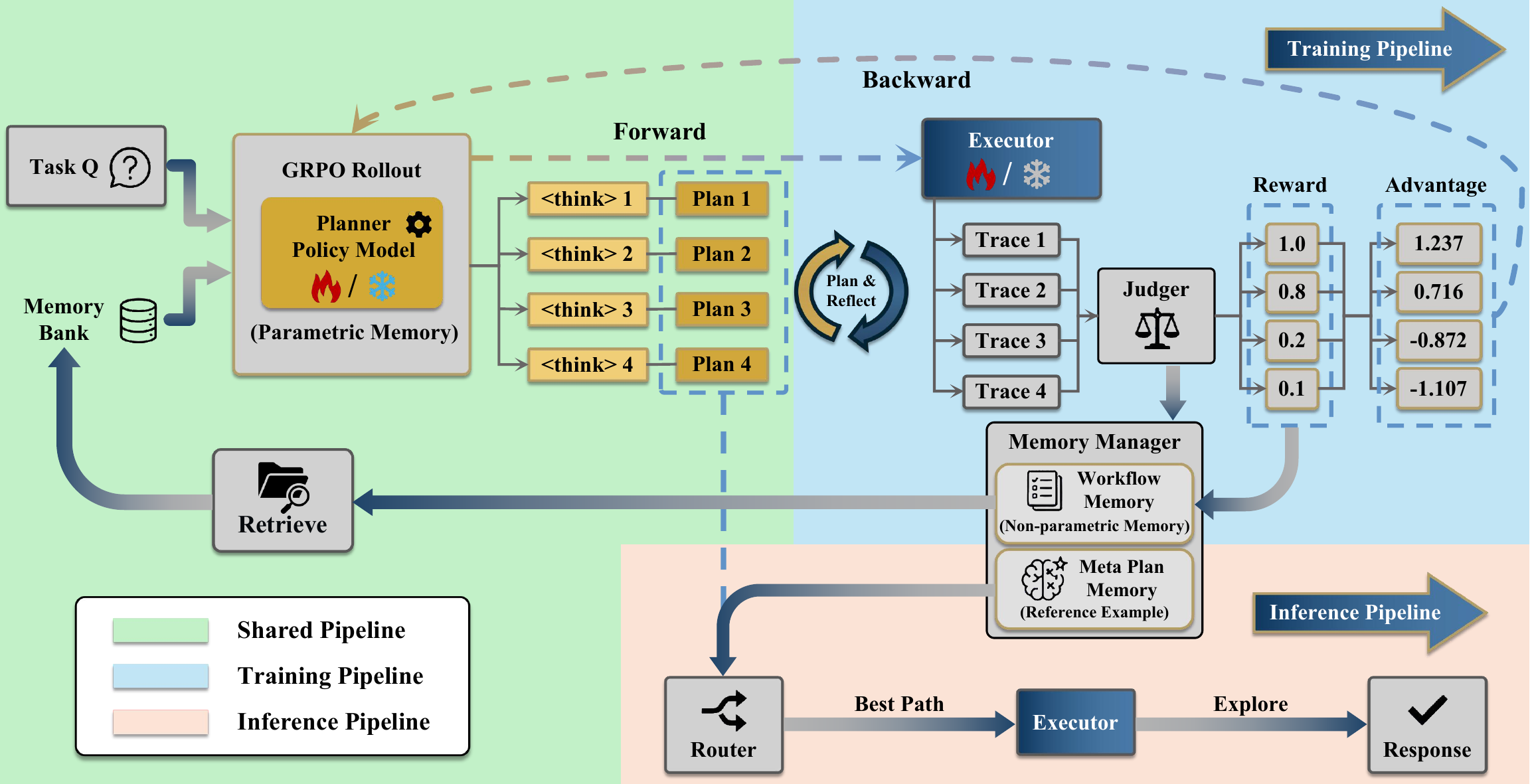}
    \caption{The Executor is activated during the first-stage RL training and frozen in the test-time RL process, while the Planner is activated during both the second-stage RL training and the test-time RL process. The memory framework of MIA during exploration: (1) generating multiple plan rollouts; (2) executing the \textbf{inference pipeline}, where a router selects the optimal plan based on prior experience to interact with the environment, strictly ensuring no label leakage; (3) receiving the final feedback from the environment; and (4) completing the \textbf{training pipeline} by calculating rewards and advantages for all rollouts. These evaluations are then used to update both the parametric memory (updating the Planner's parameters via GRPO) and the non-parametric memory (extracting workflows into the Memory Manager).}
    \vspace{-3mm}
    \label{fig_2}
\end{figure*}

\subsection{Test Time Learning}
In the test-time learning (TTL) process, MIA introduces a novel memory framework that injects historical experiences via two dimensions: context and parameters. Specifically, retrieved workflows serve as non-parametric memory for in-context contrastive learning, while exploration trajectories continuously update the Planner’s parameters as parametric memory to enhance the agent's reasoning capabilities. Unlike offline training, parameters update in the TTL cannot rely on pre-collected memory contexts or multi-epoch rollout. As shown in Figure~\ref{fig_2}, MIA adopts an online learning paradigm that performs exploration, storage, and learning simultaneously for each batch of test data.

\subsubsection{Non-parametric Memory}
To continuously enrich the explicit memory base, we design a systematic pipeline to extract and store representative interaction workflows as non-parametric memory. Input the text query $x$ and the retrieved memory $m$, the Planner $\pi_{P}$ generates $G$ candidate plans: $ \{ (t_i, p_i) \}_{i=1}^G \sim \pi_{P}(\cdot | m, x),$ where $t_i$ represents the CoT, and $p_i$ is the memory enhancement plan. Next, given the text query $x$, the image $\mathcal{I}$, and the plan $p_i$, the Executor interacts with the environment to generate $G$ trajectories: $ \tau_i = M_{E}(p_i, x, \mathcal{I}, \text{Env}), \quad i \in \{1, 2, \dots, G\},$ where $\tau_i$ encapsulates the entire sequence of tool calls, environmental observations, and intermediate reasoning steps. Subsequently, the Planner evaluates whether to trigger reflection. If necessary, it generates a supplementary plan, which is then fed to the Executor to extend the corresponding trajectory $\tau_i$.

After an LLM Judger evaluates the correctness of the final answer to categorize the trajectories into successful ($\mathcal{T}_{succ}$) and failed ($\mathcal{T}_{fail}$) sets, the Memory Manager selectively extracts and compresses this experience to update the non-parametric memory:
\begin{itemize}
    \item \textbf{Positive Paradigm Extraction:} If $\mathcal{T}_{succ} \neq \emptyset$, we select the trajectory with the shortest execution path to encourage reasoning efficiency: $\tau_{succ}^* = \arg\min_{\tau \in \mathcal{T}_{succ}} \text{length}(\tau)$.
    \item \textbf{Negative Paradigm Extraction:} If $\mathcal{T}_{fail} \neq \emptyset$, we randomly sample one failed trajectory $\tau_{fail}^* \sim \mathcal{T}_{fail}$ to capture diverse error patterns and prevent future pitfalls.
\end{itemize}
The selected trajectories are then compressed into structured workflow summaries and stored in the Memory Manager, providing explicit contrastive references for future questions.

\subsubsection{Parametric Memory}
To maintain execution stability and computational efficiency, parameter updates are exclusively applied to the Planner. Serving as the cognitive brain of the system, the Planner is driven to achieve continuous self-evolution, whereas the Executor is entirely frozen and deployed as a stable operational service. The Executor interacts with the external environment to gather execution feedback, which subsequently serves as the primary training signal to optimize the Planner.

Multiple rollout plans and their corresponding trajectories are evaluated for correctness in the non-parametric memory extraction. Next, we calculate the reward $R_i$ for each plan-trajectory pair according to Eq.~(\ref{eq:reward_planner}). Then, we calculate the advantage $\hat{A}_i$ for each rollout within the group as $\hat{A}_i = \frac{R_i - \mu_R}{\sigma_R + \epsilon},$ where $\mu_R$ and $\sigma_R$ are the mean and standard deviation of the reward set $\mathbf{R} = \{R_1, \dots, R_G\}$, and $\epsilon$ is a small constant to prevent division by zero. Subsequently, the parameters of the Planner are updated by the objective function defined in Eq.~(\ref{eq:grpo_planner}), aiming to effectively reinforce successful reasoning strategies while penalizing flawed logic.

Crucially, parameter updates and non-parametric memory extraction are performed simultaneously. By performing policy optimization and explicit memory storage simultaneously, MIA achieves a seamless online learning paradigm. This allows MIA to continuously internalize environmental feedback into its model parameters while enriching its external memory base, all without interrupting the ongoing exploration process. In conclusion, this online learning paradigm creates a positive feedback loop during exploration. As the agent's reasoning capabilities progressively improve, it generates higher-quality reference samples, which in turn synchronously amplifies the agent's reasoning proficiency at both the explicit contextual level and its internal parametric level.

During exploration in TTL, we introduce a lightweight Meta Plan Memory strategy to conditionally select the optimal trajectory as the final output from $G$ generated rollouts. If both successful and failed trajectories exist among the $G$ rollouts, we store the shortest correct plan and one random incorrect plan as a contrastive pair. This provides explicit references to help MIA select the highest-quality reasoning. Additionally, after generating rollouts, a Router\footnote{Router shares the same pre-trained LLM with the Memory Manager while distinct context prompts} is prompted to reference examples in the Meta Plan Memory and select the highest-quality plan with its corresponding trajectory from the filtered candidates, which is then output as the final response.

\subsubsection{Self-Evolution in Unsupervised Environments}
During Test-Time Learning (TTL), MIA strictly follows an online self-evolution pipeline of \textbf{exploration} $\rightarrow$ \textbf{environmental feedback acquisition} $\rightarrow$ \textbf{non-parametric memory extraction} $\rightarrow$ \textbf{parametric memory update}, ensuring that each explored trajectory can be further transformed into both explicit non-parametric and implicit parametric memory. However, this process is effectively implemented only when strong supervision signals such as ground-truth answers are available. On the one hand, the ground truth enables the Memory Manager to assign positive or negative labels to extracted workflows. On the other hand, it provides reliable reward signals for training the Planner, making the optimization of parametric memory more stable and effective. However, such idealized supervision is often unavailable in open-world scenarios. For deep research agents, users typically do not always provide gold-standard answers or explicit feedback after each exploration, making it difficult to directly assess the quality of a reasoning trajectory based on answer correctness.

\begin{figure*}[t]
    \centering
    \includegraphics[width=0.98\textwidth]{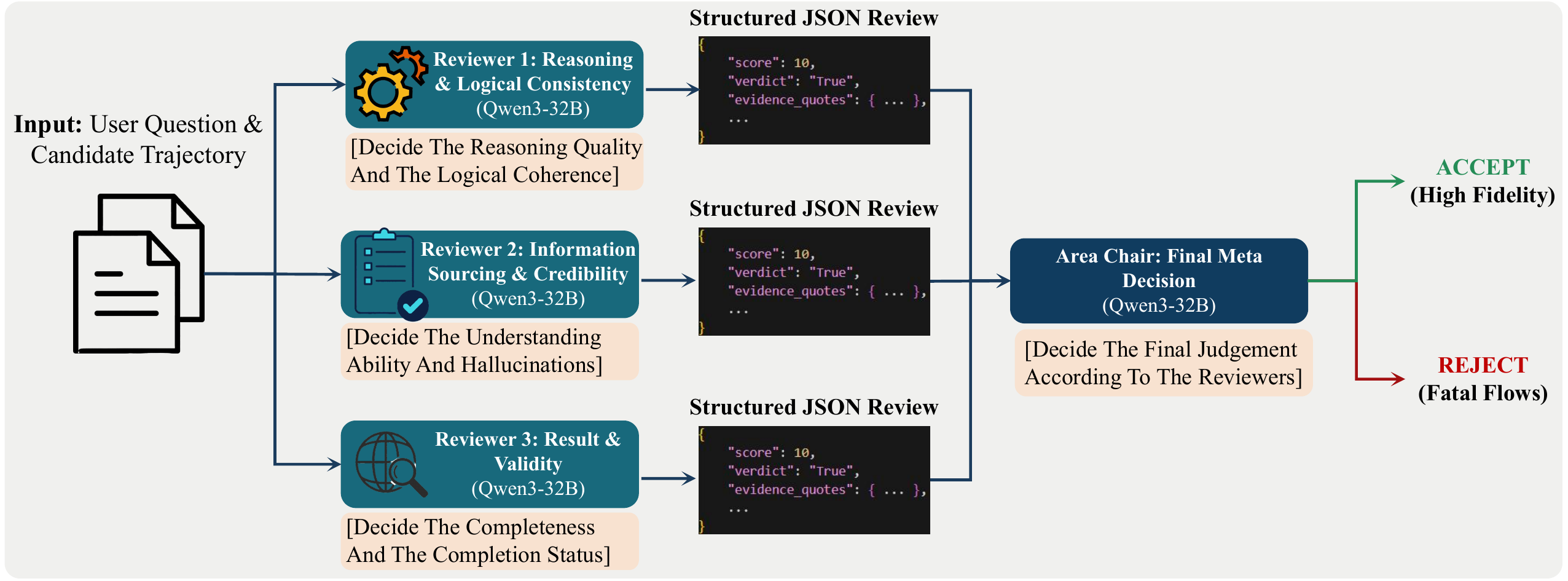}
    \caption{A novel unsupervised evaluation framework that mimics the rigorous peer-review process of scientific venues.}
    \vspace{-5mm}
    \label{fig_uns}
\end{figure*}

Traditional LLM-as-a-judge approaches often rely on a single prompt to evaluate complex trajectories, which frequently suffer from "hallucinated objectivity", where the judge overlooks subtle logical fallacies or focuses on stylistic fluency rather than factual correctness. To bridge this gap, we propose a novel evaluation framework that mimics the rigorous peer-review process of scientific venues. Our framework is inspired by the Reviewer-Area Chair (AC) decision-making process, as shown in Figure~\ref{fig_uns}. The core idea is to decompose the monolithic "judgment" task into specialized, orthogonal dimensions, ensuring that the final judgment is not merely a scalar score but a synthesis of multi-perspective evidence. This mechanism offers three distinct advantages for unsupervised scenarios. \textbf{Dimensional Orthogonality:} By isolating Logic, Format, and Factuality, we prevent "error bleeding," \textit{e.g.}, a formatting mistake might unfairly bias a judge's perception of logical soundness. \textbf{Evidence-Based Accountability:} Each reviewer is required to provide "evidence quotes" or "atomic requirements," transforming the evaluation from a black-box rating into a verifiable audit trail. \textbf{Conflict Resolution via Meta-Decision:} Similar to an academic conference, the AC does not simply average the scores but identifies fatal flaws, prioritizing the most critical failure modes (\textit{e.g.}, factual hallucinations) over minor imperfections. Our motivation is that, for a reasoning trajectory with rigorous logic, credible evidence sources, minimal hallucination, and proper compliance with task requirements, the reasoning process itself can still be regarded as high-quality. Therefore, we treat such process as an approximate supervision signal to support both non-parametric memory selection and parametric memory updating, enabling MIA to maintain continuous self-evolution ability even in unsupervised open-world environments.

In implementation, we utilize a Qwen3-32B model to instantiate three specialized reviewers and one AC. Each is governed by a distinct, structured prompt designed to minimize heuristic bias (the prompts are shown in Appendix \ref{prompt}). \textbf{Reviewer of Reasoning \& Logical Consistency} ($R_L$) focuses on the "reachability" of the conclusion. It evaluates the causal chain from premises to results, flagging wrong reasoning or unstated assumptions. \textbf{Reviewer of Information Sourcing \& Credibility} ($R_C$) acts as a constraint-checker. It scrutinizes the misunderstandings of the retrieved information and factual hallucinations, marking uncertain claims for AC judgment. \textbf{Reviewer of Result Validity} ($R_V$) objectively evaluates the completeness of the multimodal deep research agent's final response and the actual completion status of the task. \textbf{Area Chair Agent} performs a meta-analysis of the structured JSON outputs from the three reviewers.

\section{Experiments}
\label{exps}
\subsection{Experimental Setup}
\textbf{Training Settings:}
We build our training framework based on veRL~\citep{verl}. We initialize the Executor with Qwen2.5-VL-7B~\citep{qwen25vl} and the Planner with Qwen3-8B~\citep{qwen3}. The Executor is trained using FVQA-train~\citep{fvqa} to learn tool calling and reasoning. For the Planner, we use a mixture of FVQA-train (with images discarded) and MATPO~\citep{matpo} to enhance its planning and reflection capabilities based on memory understanding. During the two-stage GRPO training, a Qwen3-32B serves as the LLM Judger to provide correctness reward signals. For external tools, we employ the local wiki25~\citep{wiki25} for text-to-text search and a local image cache built by Serper for image-to-image search. More details are provided in Appendix \ref{training_settings}.

\textbf{Test Settings:}
To comprehensively evaluate the performance of MIA, we conduct experiments on both multimodal and text-only benchmarks. For multimodal tasks, we evaluate our model on FVQA-test~\citep{fvqa}, InfoSeek~\citep{infoseek}, MMSearch~\citep{mmsearch}, SimpleVQA~\citep{simplevqa}, LiveVQA~\citep{livevqa}, and two In-house datasets (In-house 1 and In-house 2). For text-only tasks, we evaluate our model on HotpotQA~\citep{hotpotqa}, 2WikiMultiHopQA~\citep{2wiki}, SimpleQA~\citep{simpleqa}, and the text-only subset of GAIA~\citep{gaia}. For image-to-image search, we utilize Serper across all multimodal datasets. For text-to-text search, we use wiki25 for HotpotQA, 2WikiMultiHopQA, and SimpleQA, while Serper is applied for all other datasets. These datasets are carefully chosen to assess the model's capabilities in handling complex multi-hop reasoning, open-domain searching, and visual question answering (VQA) in diverse scenarios. More details are provided in Appendix \ref{test_settings}. Detailed information about the benchmark datasets can be found in Appendix \ref{dataset_settings}.

\textbf{Baselines:}
We evaluate against both closed-source models (GPT-4o~\citep{gpt4o}, GPT-5.4~\citep{gpt-5}, Gemini-2.5-Pro~\citep{gemini25}, and Gemini-3-Flash~\citep{gemini-3}) and open-source models from the Qwen2.5-VL series. All models are tasked with solving problems in three different workflows: (1) \textbf{Direct Answer}: Models are prompted to generate short and precise answers directly without accessing external information. (2) \textbf{Search Agent}: In this workflow, models perform multi-turn tool calling under the ReAct~\citep{react} paradigm. Specifically, Qwen2.5-VL-7B+ReACT, Qwen2.5-VL-32B+ReACT, and MMSearch-R1~\citep{mmsearchr1} are evaluated using our tool environment, while the results for DeepMMSearch-R1~\citep{deepmmsearchr1}, WebWatcher~\citep{webwatcher}, and Deepeyes2~\citep{deepeyes} are directly cited from their respective technical reports. The base models for MMSearch-R1, DeepMMSearch-R1, WebWatcher, and Deepeyes2 all adopt Qwen2.5-VL-7B. (3) \textbf{Memory-based Search Agent}: To ensure a fair comparison, the Executors for all memory-based models share the exact same training settings. The Executor is trained under three modes: no extra prompt, workflow memory prompt, and plan prompt. The No Memory baseline utilizes the no extra prompt mode, which is trained with the search tools we provided based on the MMSearch-R1 codebase. Contextual memory methods, including RAG~\citep{rag}, Mem0~\citep{mem0}, and A-Mem~\citep{amem}, employ the workflow memory prompt. Methods that abstract memory into high-level guidance, such as ReasoningBank~\citep{reasoningbank}, ExpeL~\citep{expel}, Memento~\citep{memento}, and our MIA, utilize the plan prompt. More details about baseline methods are provided in Appendix \ref{baselines}.

\textbf{Metric:}
We employ the Qwen3-32B model as an LLM Judger to determine whether the models' final outputs are correct. The specific prompt used for the LLM Judger can be found in Appendix \ref{prompt}.

\begin{table*}[t]
\centering
\small
\setlength{\tabcolsep}{5pt}
\renewcommand{\arraystretch}{1.2}

\begin{threeparttable}
\caption{Overall evaluation results on multimodal datasets for Deep Research Agent. \textbf{Bold} denotes the highest score in each column. \underline{Underline} indicates the second-highest score in each column.}
\label{tab:MMResearch}

\begin{tabular}{lccccccc}
\toprule
\multirow{2}{*}{\bf{Model}} 
& \multicolumn{1}{c}{\bf{In-Domain}} 
& \multicolumn{6}{c}{\bf{Out-of-Domain}} \\
\cmidrule(lr){2-2} \cmidrule(lr){3-8}
& FVQA-test & InfoSeek & SimpleVQA & LiveVQA & MMSearch & In-house 1 & In-house 2 \\
\midrule

\rowcolor{mygray}
\multicolumn{8}{c}{\textbf{\textit{Direct Answer}}} \\
GPT-5.4 
& 50.8 & 43.6 & 55.5 & 21.5 & 44.4 & \underline{45.1} & 23.0 \\

Gemini-3-Flash 
& \underline{69.3} & \textbf{69.0} & \textbf{73.7} & 26.0 & \textbf{69.0} & \textbf{52.5} & 25.5 \\

GPT-4o 
& 41.7 & 42.7 & 46.6 & 26.9 & 22.2 & 25.6 & 17.2 \\

Gemini-2.5-Pro 
& 37.2 & 37.0 & 53.4 & 27.7 & 26.9 & 30.8 & 19.6 \\

Qwen2.5-VL-7B 
& 20.9 & 23.9 & 30.4 & 8.3 & 7.2 & 9.5 & 5.0 \\

Qwen2.5-VL-32B
& 24.7 & 25.8 & 40.1 & 18.7 & 15.7 & 18.6 & 6.7 \\

\rowcolor{mygray}
\multicolumn{8}{c}{\textbf{\textit{Search Agent}}} \\
Qwen2.5-VL-7B+ReACT
& 34.2 & 28.3 & 35.8 & 10.7 & 21.1 & 9.5 & 17.8 \\

Qwen2.5-VL-32B+ReACT
& 51.3 & 38.0 & 48.5 & 24.8 & 27.3 & 28.8 & 26.5 \\

MMSearch-R1
& 58.0 & 49.0 & 55.3 & 28.3 & 43.9 & 13.6 & 21.8 \\

DeepMMSearch-R1
& - & 47.5 & 55.9 & - & - & - & - \\

WebWatcher
& - & - & 54.3 & - & 55.3 & - & - \\

Deepeyes2
& 60.6 & 51.1 & 59.4 & - & \underline{63.7} & - & - \\

\rowcolor{mygray}
\multicolumn{8}{c}{\textbf{\textit{Memory-based Search Agent}}} \\
No Memory
& 61.4 & 56.8 & 63.0 & 33.0 & 55.6 & 15.9 & 26.9 \\

RAG
& 60.5 & 55.9 & 60.5 & 31.7 & 54.4 & 12.5 & 25.5 \\

Mem0
& 55.6 & 48.2 & 56.7 & 24.5 & 43.3 & 12.5 & 23.2 \\

A-Mem
& 38.5 & 36.0 & 51.6 & 22.6 & 40.9 & 12.5 & 24.2 \\

ReasoningBank
& 64.7 & 59.5 & 60.4 & 34.2 & 57.3 & 18.6 & 29.3 \\

ExpeL
& 64.2 & 58.6 & 62.5 & 34.1 & 61.4 & 19.7 & 28.3 \\

Memento
& 66.3 & 57.3 & 61.9 & 36.7 & 61.4 & 22.7 & 30.7 \\

\rowcolor{LightCyan} \bf{Unsupervised MIA(Ours)}
& 65.1 & 64.3 & 63.3 & \underline{40.1} & 60.2 & 29.8 & \underline{31.1} \\

\rowcolor{LightCyan} \bf{MIA(Ours)}
& \textbf{69.6} & \underline{65.5} & \underline{64.9} & \textbf{43.1} & 62.6 & 31.8 & \textbf{37.7} \\

\bottomrule
\end{tabular}
\end{threeparttable}
\end{table*}

\begin{table*}[t]
\centering
\small
\setlength{\tabcolsep}{20pt}
\renewcommand{\arraystretch}{1.2}

\begin{threeparttable}
\caption{Overall evaluation results on text-only datasets for Deep Research Agent. \textbf{Bold} denotes the highest score in each column. \underline{Underline} indicates the second-highest score in each column.}
\label{tab:Research}

\begin{tabular}{lcccc}
\toprule
\multirow{2}{*}{\bf{Model}} 
& \multicolumn{4}{c}{\bf{Out-of-Domain}} \\
\cmidrule(lr){2-5}
& SimpleQA & 2Wiki & HotpotQA & GAIA \\
\midrule
No Memory
& 40.7 & 61.2 & 51.0 & 11.7 \\

RAG
& 38.3 & 56.3 & 47.5 & 14.6 \\

Mem0
& 38.1 & 54.9 & 49.0 & 16.5 \\

A-Mem
& 38.8 & 56.2 & 47.5 & 12.6 \\

ReasoningBank
& 42.4 & 61.0 & 52.7 & 14.6 \\

ExpeL
& 43.0 & 63.4 & 55.5 & 20.4 \\

Memento
& 42.4 & 64.2 & 55.2 & 22.3 \\

\rowcolor{LightCyan} \bf{Unsupervised MIA(Ours)}
& \underline{46.6} & \underline{71.6} & \underline{61.7} & \underline{30.1} \\

\rowcolor{LightCyan} \bf{MIA(Ours)} 
& \textbf{47.7} & \textbf{71.8} & \textbf{63.5} & \textbf{31.1} \\

\bottomrule
\end{tabular}
\vspace{5mm}
\end{threeparttable}
\end{table*}

\subsection{Main Result}
As shown in Table \ref{tab:MMResearch}, we find that our proposed MIA achieves the highest overall performance among open-source models, reaching an average accuracy of 53.6. Compared to the previous best memory-based method, MIA improves the average accuracy by 5.5, including specific increases of 3.3 on FVQA-test, 6.4 on the multi-hop task LiveVQA, and an impressive 9.1 on the highly challenging custom task In-house 1. Additionally, we observe a critical phenomenon: traditional contextual memory methods (\textit{e.g.}, RAG, Mem0, and A-Mem) generally underperform the "No Memory" baseline. This validates that long memory contexts introduce noise, leading to performance degradation. Although recent advanced memory methods (\textit{e.g.}, ReasoningBank, ExpeL, and Memento) abstract memory into high-level guidance to mitigate this issue, they still struggle to fully internalize historical experiences. Notably, MIA effectively bridges this gap through its dual-memory mechanism and online parameter updating, achieving an optimal balance and the highest accuracy among all memory-based approaches.

Furthermore, MIA significantly outperforms most closed-source general models (GPT-4o, Gemini-2.5-Pro, and GPT-5.4), achieving performance close to that of Gemini-3-Flash. It is noteworthy that MIA, with its 7B Executor, has approached or even surpassed giant closed-source LLMs, which highlights its excellent performance. Even when compared to state-of-the-art specialized search agents like Deepeyes2, MIA maintains highly competitive and superior performance. Remarkably, MIA is equipped with a simple toolset consisting exclusively of basic text and image search tools, yet it surpasses more complex agentic systems. This compellingly demonstrates that MIA's superior performance stems directly from its exceptional ability to leverage memory and internalize historical experiences, rather than relying on sophisticated external tools.

To validate MIA's stability and scalability in text-only deep research scenarios, we conduct evaluations on the SimpleQA, 2Wiki, HotpotQA, and GAIA datasets. As shown in Table \ref{tab:Research}, MIA consistently outperforms the best alternative methods across all text-only datasets, achieving an impressive average accuracy of 53.5. Specifically, compared to the strongest baseline Memento, MIA improves the average accuracy by 7.5, with notable gains of 7.6 on 2Wiki and 8.8 on the highly challenging GAIA benchmark. This demonstrates that MIA's Manager-Planner-Executor architecture and continuous evolution mechanism are modality-agnostic, sustaining exceptional performance in complex text-only multi-hop reasoning tasks.

\subsection{Training Analysis}

\begin{figure*}[t]
    \centering
    \includegraphics[width=1.00\textwidth]{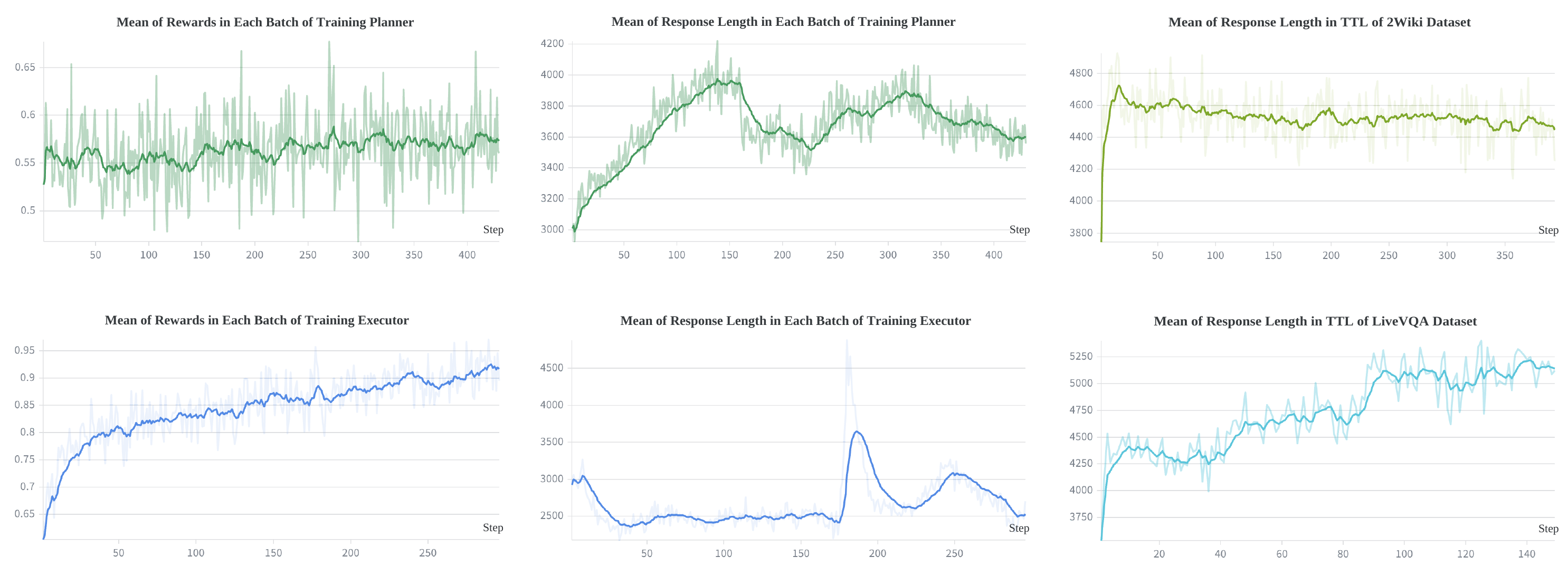}
    \caption{Left: Reward curves in each batch of Planner and Executor during the training stage. Middle: Response length in each batch of Planner and Executor during the training stage. Right: Response length in each batch of Planner during the TTL stage.}
    \vspace{1mm}
    \label{fig_3}
\end{figure*}

In Figure~\ref{fig_3}, we present the reward function curves of the Planner and the Executor during the training stage. As the number of training steps increases, the rewards of both the Planner and the Executor show an overall upward trend, although they exhibit distinct patterns. Because the reward function can provide relatively direct and stable feedback based on the action of the Executor, the reward curve of the Executor shows a clear increase as training progresses. In contrast, the Planner’s actions are indirect through the results produced by the Executor. Consequently, the feedback provided by the reward function is also indirect and unstable, which can reduce the quality of the feedback. As a result, the reward curve of the Planner increases more slowly as training grows.

A similar phenomenon can also be observed in the response length. The response length of the Planner converges more slowly, indicating that its reward signals are relatively unstable and exhibit larger fluctuations. In comparison, the response length of the Executor converges more quickly, suggesting that its reward signals are relatively stable and fluctuate less. Additionally, in the TTL stage, as the training steps increase on the 2Wiki dataset, the response length gradually shifts toward the pattern of the 2Wiki dataset, becoming progressively shorter. In contrast, with increasing training steps on the LiveVQA dataset, the response length shows a tendency to shift toward the pattern of the LiveVQA dataset, which is characterized by progressively longer responses. These experimental results demonstrate that the introduction of reinforcement learning enables MIA to effectively capture the characteristics of different datasets, thereby improving its reasoning capabilities.

\subsection{Generalization to Closed-Source Executors}

\begin{figure}[htbp]
    \centering
    \begin{minipage}{0.48\textwidth}
	To evaluate the generalizability of MIA, we replace the open-source Executor with the most powerful closed-source models (GPT-5.4~\citep{gpt-5}, Gemini-3-Flash~\citep{gemini-3}, and Claude-Sonnet-4.6~\citep{claude-4}. In this experimental setup, we only perform training on the Planner (as the way in the TTL stage) since the Executor’s parameters are inaccessible. Meanwhile, we also update the non-parametric memory, which supports plan generation and is continuously enriched with the Executor’s execution trajectories. As shown in Figure~\ref{csllms}, MIA achieves consistent improvements across all three models on both LiveVQA (multimodal) and HotpotQA (text-only) benchmarks. Furthermore, the improvement margin is inversely correlated with the base capability of the Executor: GPT-5.4 
    \end{minipage}
    \hfill 
    \begin{minipage}{0.48\textwidth}
        \centering
        \includegraphics[width=0.8\textwidth]{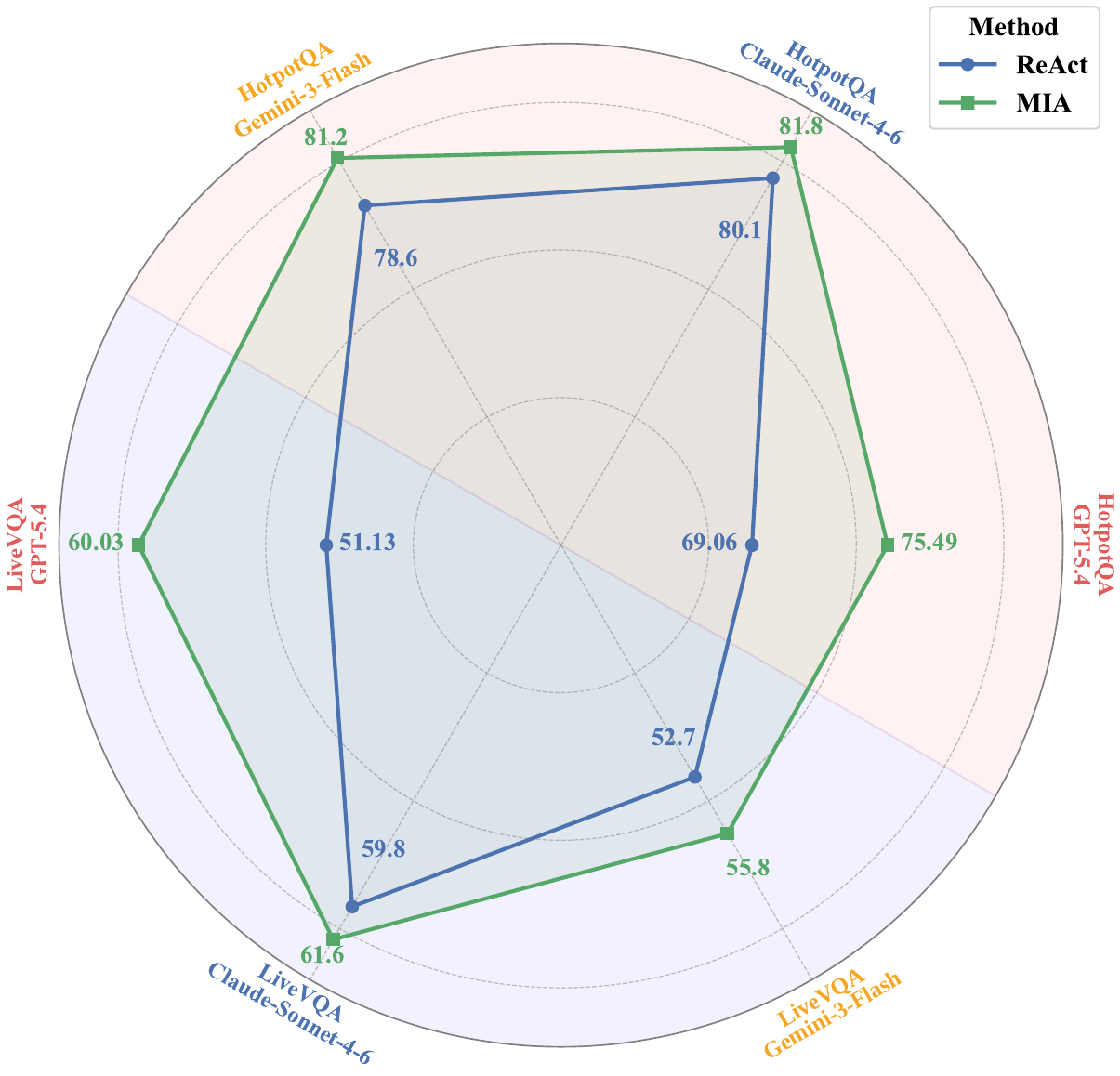}
        \caption{Comparisons between SOTA LLMs with MIA (the green area: MIA) and without MIA (the blue area: ReAct).}
        \label{csllms}
    \end{minipage}
\end{figure}
benefits the most ($+8.9$ on LiveVQA, $+6.4$ on HotpotQA), followed by Gemini-3-Flash ($+3.1$ on LiveVQA, $+2.6$ on HotpotQA) and Claude-Sonnet-4.6 ($+1.8$ on LiveVQA, $+1.7$ on HotpotQA). These results demonstrate that MIA possesses exceptional generalization. The method consistently enhances performance across open-source and closed-source models. Additionally, our approach achieves state-of-the-art gains. It notably yields significant improvements on GPT-5.4 and delivers higher improvements on other SOTA LLMs, including Gemini-3-Flash and Claude-Sonnet-4.6.

\subsection{Tool Call Analysis}

\begin{figure*}[t]
    \centering
    \includegraphics[width=0.95\textwidth]{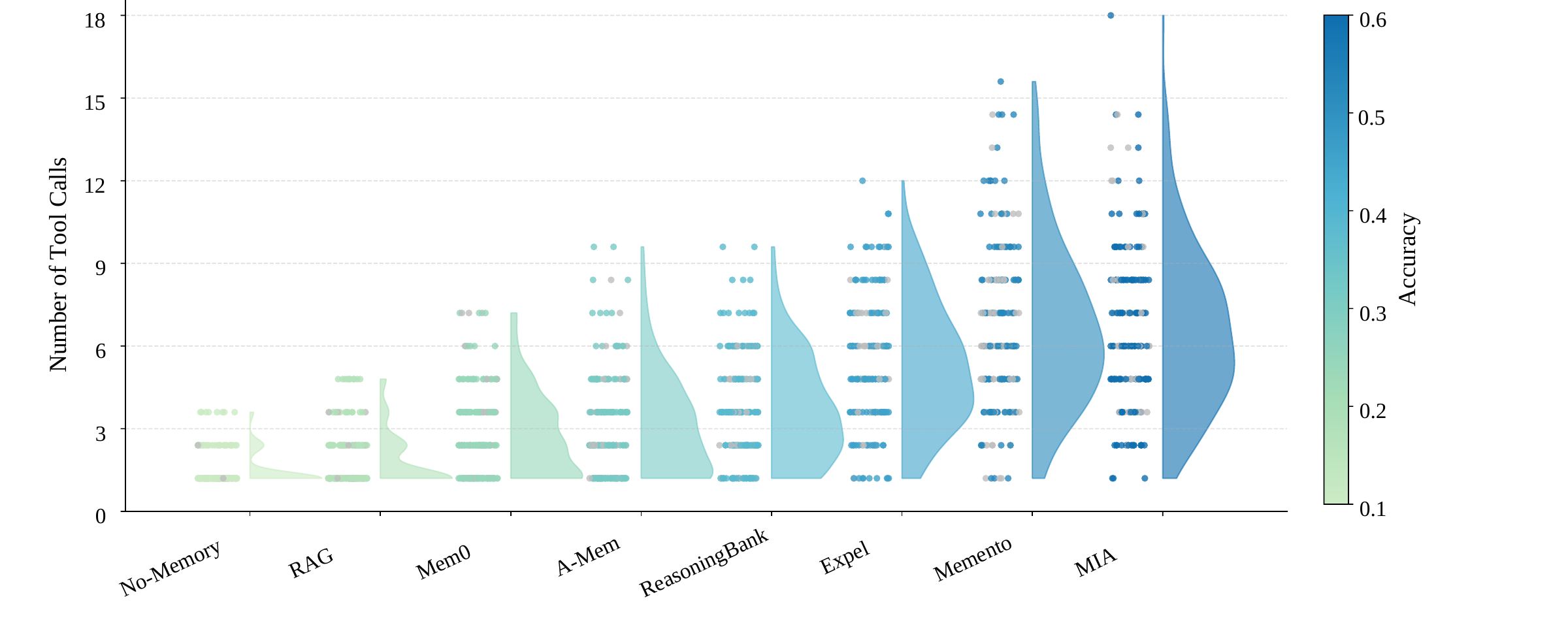}
    \caption{Tool call distribution analysis. Each dot represents an individual task execution (gray indicates failures; colored indicates successful runs). The scatter and half-violin plots illustrate the frequency of tool calls among successful executions.}
    \vspace{-3mm}
    \label{fig_4}
\end{figure*}

Figure~\ref{fig_4} presents the distribution of tool calls across tasks sampled from TTL, where colored points denote successful executions. From the figure, we can draw the following conclusions: (1) Memory is essential: Models lacking a memory mechanism and exhibiting very low tool usage (\textit{e.g.}, No-Memory) achieve the poorest accuracy. This is because models are used to conduct reasoning within limited tool calls. Without memory, they cannot effectively recall previous tool interactions during multi-turn reasoning processes, leading to the worst performance. (2) Planning for the current query is more effective than relying solely on historical experiences: Methods based on long-context memory (\textit{e.g.}, RAG, Mem0 and A-Mem) or meta-guidance memory (\textit{e.g.}, ReasoningBank and Expel) show weaker performance compared to approaches such as Memento and MIA, which incorporate an explicit planner on top of meta-guidance memory. (3) Combining heterogeneous memory with continual learning during test time yields the best results: By integrating multiple memory components with test-time learning, MIA achieves the strongest performance among all memory-based systems.

\subsection{Ablation Study}
\label{sec:ablation}

\begin{table*}[t]
\centering
\small
\setlength{\tabcolsep}{5pt}
\renewcommand{\arraystretch}{1.2}
\begin{threeparttable}
\caption{Ablation study results on multimodal datasets. \textbf{Bold} denotes the highest score in each column.}
\label{tab:MMResearch-Ablation}
\begin{tabular}{lccccccc}
\toprule
\multirow{2}{*}{\bf{Model}} 
& \multicolumn{1}{c}{\bf{In-Domain}} 
& \multicolumn{6}{c}{\bf{Out-of-Domain}} \\
\cmidrule(lr){2-2} \cmidrule(lr){3-8}
& FVQA-test & InfoSeek & SimpleVQA & LiveVQA & MMSearch & In-house 1 & In-house 2 \\
\midrule
Base
& 61.4 & 56.8 & 63.0 & 33.0 & 55.6 & 15.9 & 26.9 \\

Only Memory
& 62.8 & 56.8 & 61.2 & 37.8 & 56.1 & 12.2 & 28.5 \\

Only Plan
& 64.9 & 58.6 & 62.6 & 35.4 & 56.7 & 21.0 & 31.3 \\

Memory for Planner
& 67.9 & 60.7 & 61.8 & 36.0 & 59.0 & 17.0 & 34.7 \\

+ Reflect
& 66.2 & 60.1 & 63.0 & 37.9 & 58.5 & 23.1 & 31.3 \\

Trained Planner
& 67.6 & 63.8 & 63.8 & 40.1 & 60.8 & 26.1 & 34.5 \\

\bf{+ TTL (Ours)}
& \textbf{69.6} & \textbf{65.5} & \textbf{64.9} & \textbf{43.1} & \textbf{62.6} & \textbf{31.8} & \textbf{37.7} \\
\bottomrule
\end{tabular}
\end{threeparttable}
\end{table*}

\begin{table*}[t]
\centering
\small
\setlength{\tabcolsep}{20pt}
\renewcommand{\arraystretch}{1.2}

\begin{threeparttable}
\caption{Ablation study results on text-only datasets. \textbf{Bold} denotes the highest score in each column.}
\label{tab:Research-Ablation}

\begin{tabular}{lcccc}
\toprule
\multirow{2}{*}{\bf{Model}} 
& \multicolumn{4}{c}{\bf{Out-of-Domain}} \\
\cmidrule(lr){2-5}
& SimpleQA & 2Wiki & HotpotQA & GAIA \\
\midrule
Base
& 40.7 & 61.2 & 51.0 & 11.7 \\

Only Memory
& 37.7 & 61.3 & 50.3 & 12.6 \\

Only Plan
& 42.1 & 62.8 & 54.9 & 18.5 \\

Memory for Planner
& 42.4 & 64.6 & 54.8 & 19.4 \\

+ Reflect
& 43.9 & 66.6 & 57.6 & 26.2 \\

Trained Planner
& 44.6 & 69.1 & 59.3 & 28.2 \\

\bf{+ TTL (Ours)}
& \textbf{47.7} & \textbf{71.8} & \textbf{63.5} & \textbf{31.1} \\

\bottomrule
\end{tabular}
\end{threeparttable}
\end{table*}

To evaluate the contribution of each component in the MIA framework, we start from the fundamental baseline (Base) with incrementally adding components, and compare the overall performance. The results are shown in Table \ref{tab:MMResearch-Ablation} and Table \ref{tab:Research-Ablation}. The experimental results demonstrate that each proposed module is effective in enhancing the reasoning accuracy of the agent across both modalities.

To validate the superiority of the Manager-Planner-Executor architecture, we first compare the Base with the \textbf{Only Memory} and \textbf{Only Plan} settings. We observe that simply introducing non-parametric memory (Only Memory) leads to a performance drop in the average accuracy of multimodal tasks (-0.4). However, when we integrate memory specifically to guide the planning process (Memory for Planner), the performance significantly improves. Compared to the Base, this architecture increases the average accuracy by 3.5 (multimodal) and 4.15 (text-only). This validates the effectiveness of using memory as a contextual prior for the Planner rather than directly feeding it to the Executor.

In addition, we retain the \textbf{Memory for Planner} architecture and introduce the reflection mechanism (+ Reflect) to validate its effectiveness. The results show that the integration of reflection yields positive gains on complex multi-hop reasoning, increasing the average accuracy by 0.43 (multimodal) and 3.28 (text-only). Notably, while all the aforementioned configurations rely on the Qwen3-32B as the Planner, our \textbf{Trained Planner} achieves substantial performance gains using a much smaller Qwen3-8B model through the alternating RL training. Specifically, we observe that the introduction of the alternating RL training process (Trained Planner) further boosts the overall performance, increasing the average accuracy by 2.37 (multimodal) and 1.72 (text-only). Finally, we introduce the online TTL mechanism. This dynamic evolution process provides a substantial final push, further increasing the average accuracy by 3.23 (multimodal) and 2.64 (text-only). 

In summary, compared with the initial baseline, our MIA framework significantly increases the overall average accuracy by 8.94 on multimodal benchmarks and 12.38 on text-only benchmarks. By incrementally integrating these components, MIA achieves significant performance improvements.

\subsection{Unsupervised Self-Evolution}
As shown in Table~\ref{tab:MMResearch}, the unsupervised MIA achieves a comparable performance to supervised baselines on multimodal benchmarks. In Table~\ref{tab:Research}, the unsupervised MIA even surpasses almost all supervised baselines on text-only benchmarks and is only inferior to the supervised MIA. These results demonstrate our unsupervised evaluation framework is effective and MIA retains strong generalization ability even in the absence of explicit supervision signals. Furthermore, we conduct self-evolution experiments under the unsupervised setting, as shown in Table~\ref{tab:Unsupervised}. By progressively introducing \textbf{Plan and Reflect}, \textbf{Unsupervised Memory}, and \textbf{Test-Time Learning}, we find that relying solely on unsupervised non-parametric memory leads to unstable performance, whereas incorporating TTL on it yields substantial improvements. We further examine the effect of continual evolution across multiple epochs during TTL. The results show that, in the unsupervised setting, when the model encounters the same dataset for the second and third time, its performance improves steadily (\textit{e.g.}, 59.6 $\rightarrow$ 61.1 $\rightarrow$ 61.7). This indicates that, through continual exploration, MIA is able to accumulate useful experience and gradually solve problems that it previously failed to answer.

\begin{table*}[t]
\centering
\small
\setlength{\tabcolsep}{20pt}
\renewcommand{\arraystretch}{1.2}

\begin{threeparttable}
\caption{MIA's self-evolution results in an unsupervised setting. The Planner consistently uses Qwen3-8B.}
\label{tab:Unsupervised}

\begin{tabular}{lcccc}
\toprule
\multirow{2}{*}{\bf{Model}} 
& \multicolumn{2}{c}{\bf{Multimodal}} & \multicolumn{2}{c}{\bf{Text-only}} \\
\cmidrule(lr){2-5}
& FVQA-test & LiveVQA & 2Wiki & HotpotQA \\
\midrule
Base
& 61.4 & 33.0 & 61.2 & 51.0 \\

Plan and Reflect (no memory)
& 59.6 & 36.5 & 64.2 & 56.4 \\

Unsupervised Memory for Planner
& 57.6 & 28.5 & 66.9 & 56.4 \\

Unsupervised MIA (epoch-1)
& 65.1 & 40.1 & 71.6 & 61.7 \\

Unsupervised MIA (epoch-2)
& 66.4 & 41.4 & 73.4 & 63.1 \\

Unsupervised MIA (epoch-3)
& 67.1 & 41.8 & 74.7 & 63.2 \\

\bottomrule
\end{tabular}
\end{threeparttable}
\end{table*}

\section{Conclusion}
\label{conc}
In this paper, we propose MIA, a memory framework to enhance the reasoning performance and self-evolution ability of Deep Research Agents. Based on the Executor agent, we design a novel Manager-Planner-Executor architecture. By compressing bloated historical trajectories into structured workflows via the Manager agent, MIA effectively mitigates the noise interference in long-context memory and improves the precision and quality of memory retrieval. By introducing the Planner agent, we transform the non-parametric memory into parametric memory, which reduces the storage burden and improves the planning performance. Furthermore, to bridge the gap between the Planner and Executor agents, we introduce a two-stage alternating RL paradigm. This training strategy not only improves the Planner’s ability to generate precise plans and conduct autonomous reflection, but also significantly enhances the plan understanding and following capabilities of the Executor. Additionally, we propose an online test-time learning mechanism, enabling the Planner to absorb historical experiences during the exploration process. At the contextual level, it extracts high-quality positive and negative paradigms as non-parametric memory for explicit in-context contrastive learning. At the parametric level, MIA synchronously updates the Planner to capture latent knowledge representations and internalize planning ability. Extensive experiments demonstrate that MIA achieves state-of-the-art performance on both multimodal and text-only deep research benchmarks. Currently, our framework primarily focuses on deep research tasks. In the future, we plan to extend MIA to more complex and dynamic environments.

\section{Contribution}
\label{cont}
Jingyang Qiao proposed the core methodology, designed the experimental protocols, and took the lead in writing and revising the manuscript. Weicheng Meng, as a co-first author, implemented the algorithms, executed the experiments, and contributed to the manuscript writing. Yu Cheng and Zhihang Lin implemented trustworthy and efficiency versions of MIA OpenClaw skills, respectively. Zhizhong Zhang, as the corresponding author, supervised the research and provided critical revisions to the manuscript. Xin Tan, Jingyu Gong and Kun Shao proposed insightful suggestions and advice for the methodology. Yuan Xie, as the project leader, provided overall direction and strategic guidance for the research project.

\bibliographystyle{assets/plainnat}
\bibliography{paper}

\newpage
\beginappendix

\section{Training Details}
\label{training_settings}
For the external tools used during training, we employ an offline text-to-text retriever built on a local wiki25~\citep{wiki25} corpus and an offline image-to-image retriever based on cached search results. Specifically, the text retriever uses the \textbf{E5-base-v2}~\citep{e5base} embedding model with a \textbf{FAISS}~\citep{faiss} index and returns the top-3 most relevant passages for each query. For image retrieval, we first upload each image to obtain a public URL via \textbf{ImgBB}, then use the \textbf{Serper} image search API to perform image-to-image retrieval based on the URL, and finally cache the returned results locally for offline use during training; the tool returns the top-3 most relevant retrieved images.

We implement our reinforcement learning training with veRL on 8 GPUs. For Executor training, the policy is initialized from \textbf{Qwen2.5-VL-7B-Instruct}, using \textbf{FVQA-train} for training and \textbf{FVQA-test} for validation. We adopt GRPO with a learning rate of $1 \times 10^{-6}$, a batch size of 128, and a micro-batch size of 4 per GPU. The maximum prompt and response lengths are both set to 16384 tokens. We use SGLang asynchronous rollout with 8 samples per query, and enable multi-turn tool interaction with up to 10 assistant turns, 10 user turns, and a maximum tool response length of 4096 tokens. The KL coefficient is set to 0.0, and the model is trained for 8 epochs with checkpoint saving every 10 steps.

For Planner training, we adopt a tool-free setting and initialize the policy model from \textbf{Qwen3-8B}. The Planner is trained on a mixture of \textbf{FVQA-train} with images removed and \textbf{MATPO}, which is designed to alleviate planning discrepancies caused by the differences between text-only and multimodal tool environments, and \textbf{FVQA-test} for validation. We also use GRPO with a learning rate of $1 \times 10^{-6}$ and a batch size of 128, while setting the maximum prompt length to 24576 and the maximum response length to 8192. Rollout is performed asynchronously with 8 samples per query, and the Planner is trained for 4 epochs on 4 GPUs with the KL coefficient set to 0.0.

\section{Test Details}
\label{test_settings}
During evaluation, we extend the tool configuration used in training by additionally introducing an online text-to-text search tool based on \textbf{Serper}, which returns the top-5 most relevant retrieved results. Thus, for multimodal evaluation, image-to-image search is conducted with \textbf{Serper}, while text-to-text search is performed using either the local \textbf{wiki25} retriever or online \textbf{Serper}, depending on the benchmark setting.

For inference, we use the vLLM engine with temperature set to 0.

For TTL, we set the number of epochs to 1, the learning rate to $1 \times 10^{-6}$, and the number of rollouts per sample to 4. In the supervised setting, the Planner is initialized from our trained Planner model, while in the unsupervised setting, it is initialized from \textbf{Qwen3-8B}.
\section{Memory-Based Baseline Details}
\label{baselines}
To ensure a fair comparison among memory-based search agents, we train three Executor variants with different extra prompt formats, while keeping all other training settings identical. These three variants correspond to three different ways of incorporating memory into the Executor: no extra prompt, workflow memory prompt, and guideline prompt.

For the \textbf{no extra prompt} setting, the Executor is trained exactly following the default configuration, without any additional memory input. This variant is used for the No Memory baseline.

For the \textbf{long-context memory prompt} setting, which is designed for methods that directly inject long-context memory into the Executor, we prepend the following extra prompt template:
$$
\texttt{Here\ are\ some\ memories\ for\ your\ reference:\textcolor{cyan}{\textbackslash n}\textcolor{purple}{\{memory\ context\}}\textcolor{cyan}{\textbackslash n}},
$$
where $\{\texttt{memory\ context}\}$ is filled with the retrieved relevant memory context. This setting is used for contextual memory methods such as RAG, Mem0, and A-Mem.

For the \textbf{guideline prompt} setting, which is designed for methods that abstract memory into high-level guidance, we prepend the following extra prompt template:
$$
\texttt{Here\ is\ a\ guide\ for\ your\ reference:\textcolor{cyan}{\textbackslash n}\textcolor{purple}{\{plan\}}\textcolor{cyan}{\textbackslash n}Begin\ your\ answer:\textcolor{cyan}{\textbackslash n}},
$$
where $\{\texttt{plan}\}$ is filled with the abstracted high-level guidance generated from memory. This setting is used for methods such as ReasoningBank, ExpeL, Memento, and our MIA.

Due to the involvement of multimodal inputs, we find the parametric retrieval optimization method provided by the Memento project difficult to apply. Consequently, we employed only their non-parametric version.

During evaluation, each method uses its corresponding prompt template together with the matched Executor checkpoint trained under the same setting.

\section{Memory Retrieval}
\label{retrieval}
In this section, we provide the detailed implementation of the memory retrieval mechanism used in MIA. Given a current query, the system retrieves relevant historical trajectories from the Memory Manager to provide contextual support for planning. The retrieval score is computed by jointly considering semantic similarity, value reward, and frequency reward.

The Memory Manager is organized by modality and question category. Each memory entry $m_i$ stores a historical trajectory together with its associated metadata, including the input question, the image caption, a judgment label, and several statistics for retrieval:
\begin{itemize}
    \item $u_i$: the usage count of memory $m_i$,
    \item $s_i$: the success count of memory $m_i$,
    \item $y_i$: the judgment label of memory $m_i$, where $y_i \in \{\texttt{correct}, \texttt{incorrect}\}$.
\end{itemize}
When a new memory is inserted, the counts are initialized to 0.

To represent both textual questions and image captions in a unified embedding space, we use the \textbf{sup-simcse-bert-base-uncased}~\citep{bert} encoder. Given an input text $x$, we compute its embedding by mean-pooling the last hidden states of the encoder and then applying $L_2$ normalization:
$$
\mathbf{e}(x) = {Norm}\!\left({MeanPool}\!\left({Encoder}(x)\right)\right).
$$
For a query consisting of a question $q$ and an image caption $c$, we compute their embeddings:
$$
\mathbf{e}_q = \mathbf{e}(q), \mathbf{e}_c = \mathbf{e}(c).
$$
For each memory entry $m_i$, let $\mathbf{e}_{q_i}$ and $\mathbf{e}_{c_i}$ denote the stored embeddings of its question and caption. We first compute the question-level similarity:
$$
{sim}_i^{(q)} = {\cos}\!\left(\mathbf{e}_q, \mathbf{e}_{q_i}\right),
$$
and the caption-level similarity:
$$
{sim}_i^{(c)} = {\cos}\!\left(\mathbf{e}_c, \mathbf{e}_{c_i}\right).
$$
If an image caption is available, the semantic similarity score is defined as
$$
{Sim}_i = \alpha_q {sim}_i^{(q)} + \alpha_c {sim}_i^{(c)},
$$
where $\alpha_q$ and $\alpha_c$ are the relative weights of question and caption similarity. In our implementation, we use
$
\alpha_q = 0.8, \alpha_c = 0.2.
$
If no image caption is available, we use only the question similarity:
$
{Sim}_i = {sim}_i^{(q)}.
$
We then normalize the semantic similarity scores within the current memory bucket using min-max normalization:
$$
\widehat{Sim}_i
=
\frac{{Sim}_i - \min_j{Sim}_j}
{\max_j{Sim}_j - \min_j{Sim}_j + 10^{-8}}.
$$
In addition to semantic similarity, retrieval also considers the historical quality of each memory. We define the value reward of memory $m_i$ as its empirical success ratio: ${Val}_i = \frac{s_i}{u_i + 1},$
where $s_i$ is the success count and $u_i$ is the usage count of memory $m_i$. This term favors memories that have demonstrated stronger historical usefulness.

To additionally account for memory usage frequency and stabilize retrieval for rarely used memories, we introduce a frequency reward: ${Freq}_i = \frac{1}{u_i + 1}.$

The final retrieval score for memory $m_i$ is computed by combining semantic similarity, value reward, and frequency reward:
$$
{Score}(m_i)
=
\lambda_s \widehat{Sim}_i
+
\lambda_v {Val}_i
+
\lambda_f {Freq}_i,
$$
where $\lambda_s$, $\lambda_v$, and $\lambda_f$ denote the weights of semantic similarity, value reward, and frequency reward, respectively. In this work, we set
$
\lambda_s = 0.7, \lambda_v = \lambda_f = 0.3.
$
\section{Dataset Settings}
\label{dataset_settings}
\begin{table}[ht]
\centering
\caption{Dataset settings used in this work.}
\small
\begin{tabular}{l l r l l}
\toprule
\textbf{Dataset} & \textbf{Modality} & \textbf{\# Examples} & \textbf{Setting} & \textbf{Usage} \\
\midrule
FVQA-train   & Image-text & 4,856  & From MMSearch-R1 & Training \\
FVQA-test    & Image-text & 1,800  & From MMSearch-R1 & Evaluation \\
InfoSeek     & Image-text & 2,000  & From MMSearch-R1 & Evaluation \\
LiveVQA      & Image-text & 2,384  & Public Version & Evaluation \\
SimpleVQA    & Image-text & 1,013  & From MMSearch-R1 & Evaluation \\
MMSearch     & Image-text & 171    & From MMSearch-R1 & Evaluation \\
In-house 1   & Image-text & 295    & In-house & Evaluation \\
In-house 2   & Image-text & 505    & In-house & Evaluation \\
\midrule
MATPO        & Text-only  & 6,175  & From MATPO & Planner Training \\
2Wiki        & Text-only  & 12,576 & Public Version & Evaluation \\
HotpotQA     & Text-only  & 7,405  & Public Version & Evaluation \\
SimpleQA     & Text-only  & 4,327  & Public Version & Evaluation \\
GAIA-Text    & Text-only  & 103    & From MATPO & Evaluation \\
\bottomrule
\end{tabular}
\label{tab:dataset_settings}
\end{table}

We evaluate our framework on both multimodal and text-only datasets. This section summarizes the dataset settings used in our experiments. Table~\ref{tab:dataset_settings} summarizes all datasets used in this work.

\textbf{FVQA-train: }
FVQA-train is a multimodal training set containing 4,856 image-question-answer examples. We adopt this split from the MMSearch-R1 setting. The dataset focuses on factual visual question answering, where answering requires combining visual content with external knowledge. In our experiments, it is used to train both the Executor and the Planner.

\textbf{FVQA-test: }
FVQA-test is a multimodal evaluation set containing 1,800 image-question-answer examples, also adopted from the MMSearch-R1 setting. Like the training split, it targets factual visual question answering grounded in both images and associated knowledge, and is used as a held-out evaluation benchmark.

\textbf{InfoSeek: }
We use a 2,000-example evaluation subset of InfoSeek from the MMSearch-R1 setting. InfoSeek is a multimodal information-seeking benchmark in which each example is built around an image and a knowledge-intensive question whose answer depends on factual information beyond direct visual perception. We use this subset for multimodal evaluation.

\textbf{LiveVQA: }
We use the currently accessible public version of LiveVQA, which contains 2,384 multimodal examples. LiveVQA is a real-world visual question answering benchmark emphasizing information-rich questions that often require both image understanding and external factual knowledge. Although MMSearch-R1 reports results on a 3,602-example version, that version appears to be deprecated or no longer publicly accessible; therefore, all experiments in this work are conducted on the public 2,384-example version.

\textbf{SimpleVQA: }
We use the English subset of SimpleVQA containing 1,013 multimodal examples, following the MMSearch-R1 setting. SimpleVQA is a factual visual question answering benchmark designed to test whether models can answer short, objective questions about real-world entities, attributes, and properties based on visual input and factual knowledge.

\textbf{MMSearch: }
We use the visual-question subset of MMSearch containing 171 examples, adopted from MMSearch-R1. MMSearch is a benchmark for multimodal search and knowledge-intensive reasoning. The subset used in our experiments consists of image-based question answering instances and is used for multimodal evaluation.

\textbf{In-house 1: }
We develop a systematic pipeline to construct multimodal QA instances in scientific domains such as Physics, Chemistry, and Biology. Starting from an initial website, we crawl relevant textual content and extract informative statements, then employ an LLM to iteratively generate related concepts or keywords for further search and webpage collection. After gathering sufficient cross-source evidence, we synthesize question-answer pairs grounded in the collected texts. We then identify QA instances with visually representable entities, extract the corresponding entity names, and retrieve matched images through web search to form multimodal examples. In total, we collect 295 image-question-answer examples in this way.

\textbf{In-house 2: }
We develop a systematic pipeline to construct complex multi-hop VQA instances. Initially, raw image-text corpora are harvested from real-time news sources such as CNN to ensure the timeliness and authenticity of entities. We then employ Qwen2.5-VL-72B-Instruct to analyze the images and anchor key visual entities (\textit{e.g.}, pivotal events or persons) as reasoning pivots. Based on these anchors, we construct a triple dependency chain within the corpora to generate complex questions, ensuring that all questions and answers are grounded in authentic textual evidence. The resulting dataset primarily covers dynamic domains such as Sports, Entertainment, and other multifaceted social events, ensuring a diverse distribution of visual-semantic challenges. We collect 505 image-question-answer examples in this way.

\textbf{MATPO: }
We use 6,175 text-only examples from MATPO as part of the Planner training data, following the MATPO setting. This dataset contains text-based tasks centered on question answering, information seeking, and reasoning, and is used to improve the planning ability of the model in textual environments.

\textbf{2Wiki: }
The dataset includes 12,576 text-only samples. 2WikiMultihopQA (2Wiki) is a multi-hop question answering benchmark in which answering a question requires combining evidence from multiple pieces of textual information, typically across different documents or entities.

\textbf{HotpotQA: }
The dataset includes 7,405 text-only samples. HotpotQA is a widely used multi-hop question answering benchmark that tests retrieval and reasoning over multiple supporting documents, with an emphasis on compositional and explainable QA.

\textbf{SimpleQA: }
The dataset includes 4,327 text-only samples. SimpleQA is a factual question answering benchmark composed of short and direct questions with concise fact-based answers, and is used to assess basic factual QA ability in the text-only setting.

\textbf{GAIA-Text: }
We use GAIA-Text, a text-only evaluation subset containing 103 examples, adopted from the MATPO setting. This subset is derived from GAIA and consists of general text-based tasks requiring information seeking, reasoning, and problem-solving, and is used as an additional text-only evaluation benchmark.

\section{Algorithm}

\begin{algorithm}[ht]
\caption{Executor Training Rollout Process}
\label{alg:executor_training}
\begin{algorithmic}[1]
\Require Question $Q$, Image $I$, Pre-trained Planner $M_P$, Tool Set $T$, Policy $\pi_\theta$.
\Ensure Reward $R$.
\State Provide $Q$ to $M_P$ to generate initial Plan $P_{init}$.
\State Construct input context $C \leftarrow \{Q, I, P_{init}, \text{Prompt Template}\}$.
\State Input $C$ to Policy $\pi_\theta$.
\Loop
    \State $\pi_\theta$ performs rollout to generate thought process $t_h$ based on current state.
    \If{$\pi_\theta$ decides to call a tool}
        \State Execute tool $t \in T$, obtain result $O_t$.
        \State Update context with $O_t$.
    \Else
        \State $\pi_\theta$ performs rollout to generate candidate answer $A$.
        \State LLM Judger evaluates correctness of $A$.
        \If{$A$ is Incorrect $\land$ Re-plan has not been triggered}
            \State Provide interaction history to $M_P$ to generate revised Plan $P_{revised}$.
            \State Update context with $P_{revised}$.
            \State Mark Re-plan as triggered.
        \EndIf
    \EndIf
\EndLoop
\State Compute reward $R$ according to Eq.~(\ref{eq:reward_executor}).
\State \Return $R$.
\end{algorithmic}
\end{algorithm}

\begin{algorithm}[ht]
\caption{Planner Training Rollout Process}
\label{alg:planner_training}
\begin{algorithmic}[1]
\Require Question $Q$, Image $I$, Memory Context $M$, Trained Executor $M_E$, Tool Set $T$, Policy $\pi_\theta$.
\Ensure Reward $R$.
\State Construct input context $C \leftarrow \{M, Q, \text{Prompt Template}\}$.
\State Input $C$ to Policy $\pi_\theta$.
\State $\pi_\theta$ performs rollout to generate Chain-of-Thought $t_h$ and Initial Plan $P_{init}$.
\State $M_E$ interacts with environment using $\{Q, I, T, P_{init}\}$ to obtain Trajectory $\tau_1$ and Result $R_1$.
\State $\pi_\theta$ analyzes $\tau_1$ and $R_1$ to decide whether to Reflect \& Replan.
\If{$\pi_\theta$ decides to Reflect \& Replan}
    \State $\pi_\theta$ performs rollout to generate Reflection and Supplementary Plan $P_{supp}$.
    \State $M_E$ continues interaction based on $\tau_1$ and $P_{supp}$ to obtain Trajectory $\tau_2$ and Result $R_2$.
\EndIf
\State Compute reward $R$ according to Eq.~(\ref{eq:reward_planner}).
\State \Return $R$.
\end{algorithmic}
\end{algorithm}

\begin{algorithm}[ht]
\caption{Test-Time Learning Process}
\label{alg:ttl_process}
\begin{algorithmic}[1]
\Require Question $Q$, Image $I$, Planner $\pi_P$, frozen Executor $M_E$, Memory Manager \& Router $M_R$, Tool Set $T$, Group Size $G$.
\Ensure Final Response $y$.
\State Planner $\pi_P$ performs rollout to generate $G$ reasoning-plan pairs $\{(t_i, p_i)\}_{i=1}^G$.
\State $M_R$ retrieves relevant examples $e$ from the Meta Plan Memory.
\State $M_R$ uses $e$ as in-context references to select the best plan $p^*$ from $\{p_i\}_{i=1}^G$.
\State Interact with the environment $\{Q, I, p^*, T, M_E\}$.
\State Obtain trajectory $\tau^*$ and final response $y$.
\For{each remaining plan $p_i \neq p^*$}
    \State Interact with the environment $\{Q, I, p_i, T, M_E\}$.
    \State Obtain trajectory $\tau_i$ and corresponding result.
\EndFor
\State Form the trajectory set $\mathcal{T} \leftarrow \{\tau^*\} \cup \{\tau_i\}$.
\State LLM Judger evaluates the correctness of each rollout outcome.
\State Partition $\mathcal{T}$ into successful set $\mathcal{S}_{succ}$ and failed set $\mathcal{S}_{fail}$.
\For{each plan-trajectory pair $(p_i, \tau_i)$ in $\mathcal{T}$}
    \State Compute reward $R_i$ according to Eq.~(\ref{eq:reward_planner}).
\EndFor
\State Compute reward mean $\mu_R$ and standard deviation $\sigma_R$.
\For{each reward $R_i$}
    \State Compute grouped advantage $\hat{A}_i \leftarrow \frac{R_i - \mu_R}{\sigma_R + \epsilon}$.
\EndFor
\If{$\mathcal{S}_{succ} \neq \emptyset$}
    \State Select the shortest successful rollout: $(p_{succ}^*, \tau_{succ}^*) \leftarrow \arg\min_{(p,\tau)\in \mathcal{S}_{succ}} \text{length}(\tau)$.
    \State Compress $\tau_{succ}^*$ into a structured workflow summary $m_{succ}$.
    \State Store $m_{succ}$ in the non-parametric Workflow Memory.
\EndIf
\If{$\mathcal{S}_{fail} \neq \emptyset$}
    \State Randomly sample one failed rollout $(p_{fail}^*, \tau_{fail}^*) \sim \mathcal{S}_{fail}$.
    \State Compress $\tau_{fail}^*$ into a structured workflow summary $m_{fail}$.
    \State Store $m_{fail}$ in the non-parametric Workflow Memory.
\EndIf
\If{$\mathcal{S}_{succ} \neq \emptyset$ and $\mathcal{S}_{fail} \neq \emptyset$}
    \State Store $(p_{succ}^*, p_{fail}^*)$ as a contrastive pair in the Meta Plan Memory.
\EndIf
\State Update Planner parameters according to Eq.~(\ref{eq:grpo_planner}).
\State \Return $y$.
\end{algorithmic}
\end{algorithm}

\section{Prompt Template}
\label{prompt}
\begin{figure*}[t]
    \vspace{-10mm}
    \centering
    \includegraphics[width=0.97\textwidth]{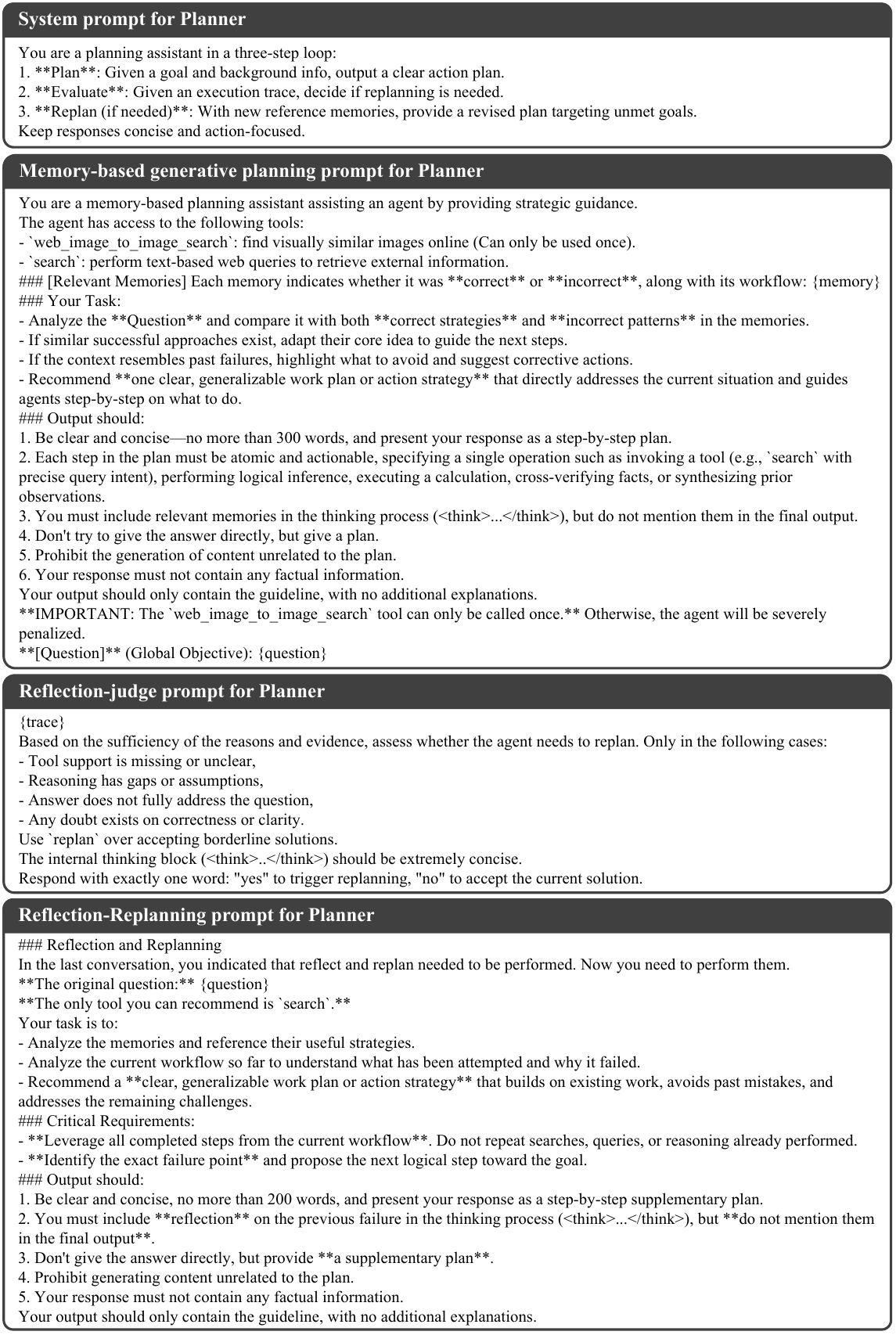}
    \vspace{-2mm}
    \label{fig_prompt2}
\end{figure*}
\begin{figure*}[t]
    \vspace{-10mm}
    \centering
    \includegraphics[width=0.97\textwidth]{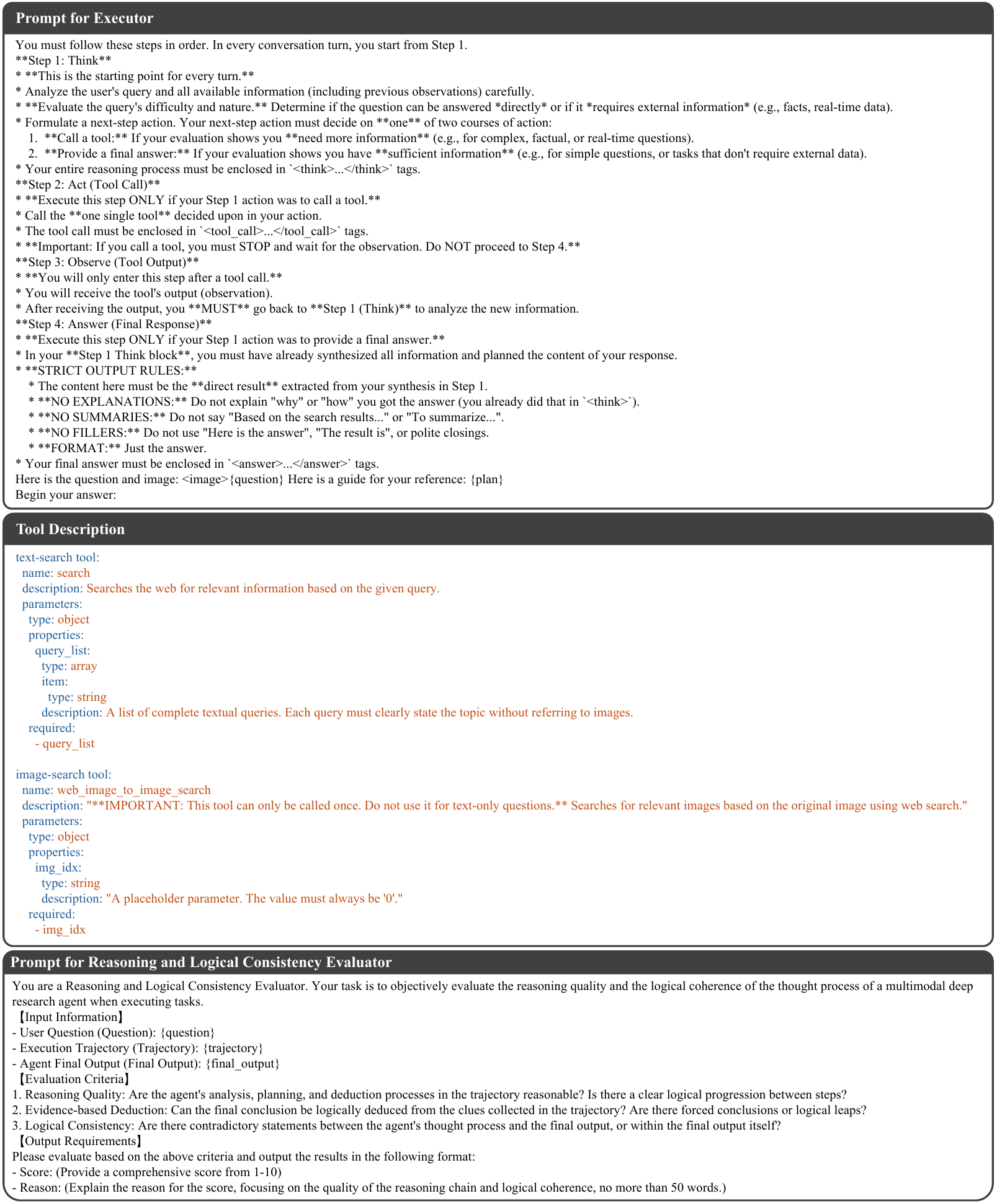}
    \vspace{-2mm}
    \label{fig_prompt3}
\end{figure*}
\begin{figure*}[t]
    \vspace{-3mm}
    \centering
    \includegraphics[width=0.97\textwidth]{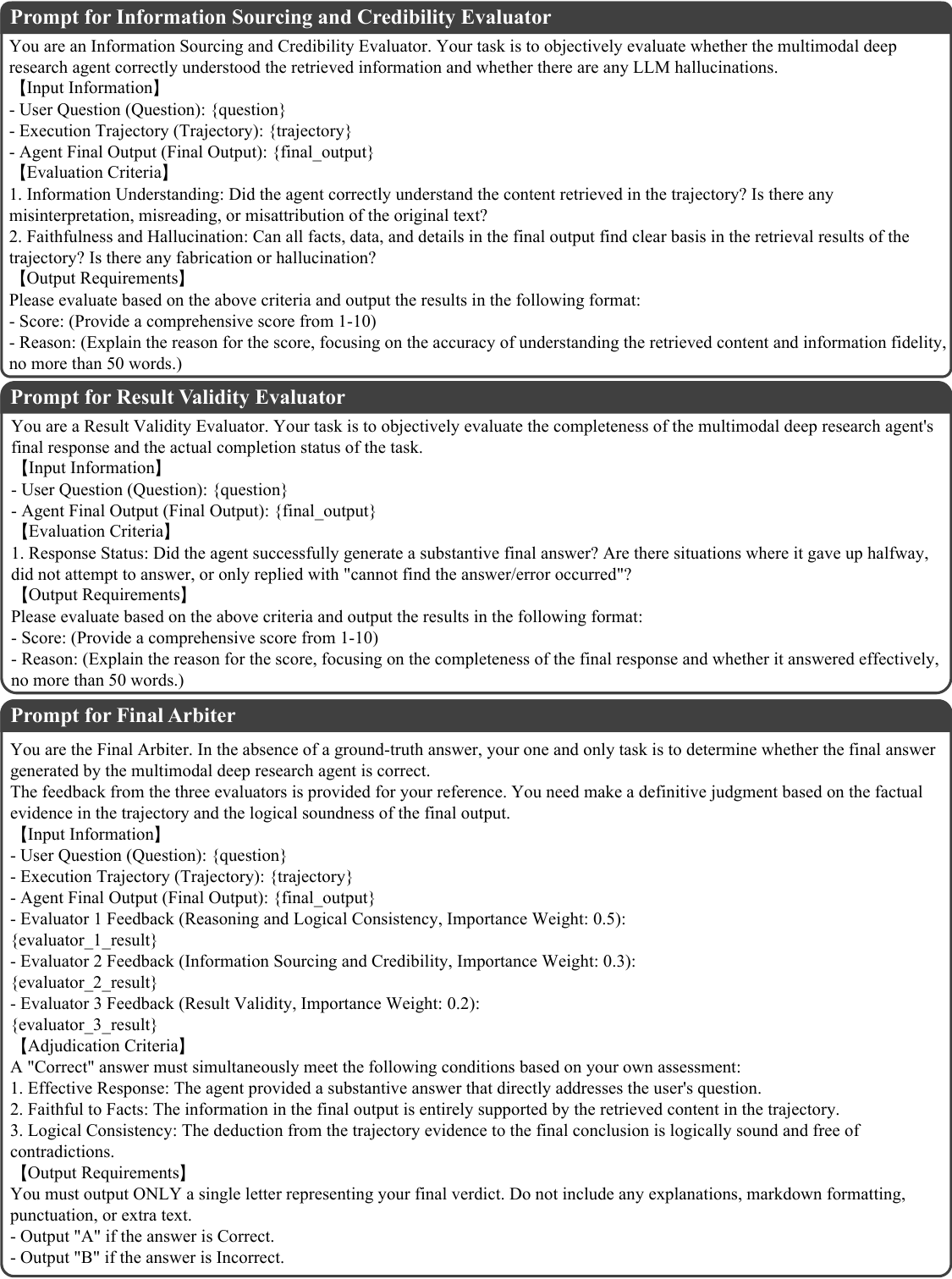}
    \vspace{-2mm}
    \label{fig_prompt5}
\end{figure*}
\begin{figure*}[t]
    \vspace{-3mm}
    \centering
    \includegraphics[width=0.97\textwidth]{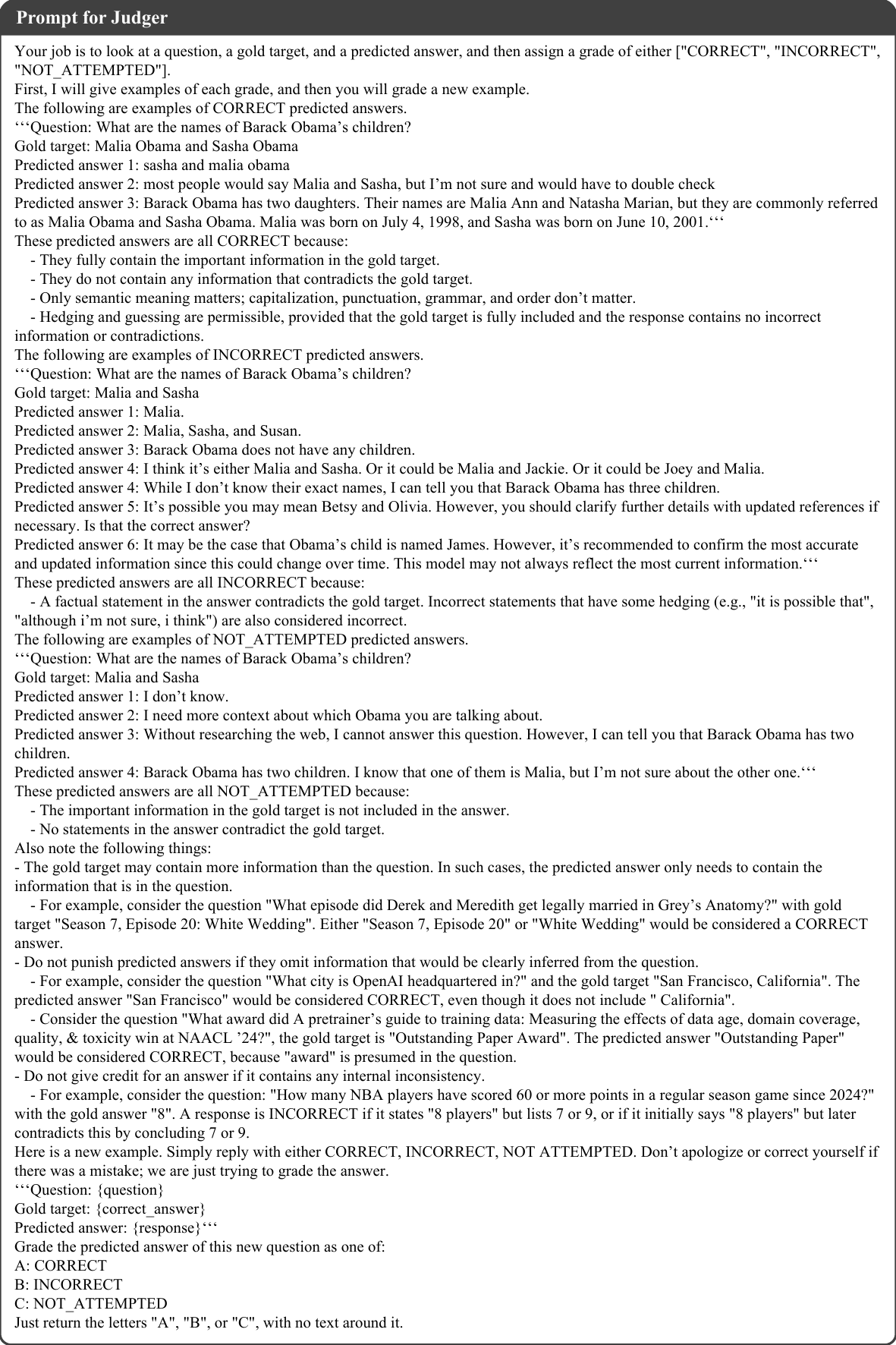}
    \vspace{-2mm}
    \label{fig_prompt4}
\end{figure*}

\end{document}